\documentclass[letterpaper]{article}
\usepackage[preprint]{aaai2027}
\usepackage[hyphens]{url}
\usepackage{graphicx}
\usepackage{natbib}
\usepackage{caption}
\usepackage{amsmath}
\usepackage{amssymb}
\usepackage{booktabs}
\usepackage{multirow}
\usepackage{dsfont}
\title{Vision-TL-Action: Neuro-Symbolic Trajectory Generation from Visual Observations and Temporal Logic}
\author{
Zezhi Liu\textsuperscript{1,2}\equalcontrib,
Zhiwei Zheng\textsuperscript{1}\equalcontrib,
Hanqian Luo\textsuperscript{1,3},
Deyun Qin\textsuperscript{1},\\
Shizhen Wu\textsuperscript{1},
Yongchun Fang\textsuperscript{1}\corresponding
}
\affiliations{\textsuperscript{1} Nankai University\\
\textsuperscript{2} Zhongguancun Academy\\
\textsuperscript{3} The Hong Kong Polytechnic University\\}

\begin{document}
\maketitle

\begin{abstract}
Temporal logic (TL) provides a compositional language for the formulation of long horizon robotic tasks, 
but existing TL-conditioned trajectory generators can sidestep perception-to-symbol binding by encoding exact object geometry in the task graph.
We introduce \emph{Vision-TL-Action}, which generates action trajectories from multi-view images, a coordinate-free TL syntax graph, and the robot initial state. TL-node tokens and spatial visual tokens are fused through bidirectional cross-attention, and the resulting representation conditions a flow-matching trajectory generator. Visual tokens are augmented only with normalized image-plane locations and camera-view identifiers, while a training-only predicate-to-region objective encourages grounding to referenced objects. Consistent with prior work in this domain, we evaluate the model using Success@$K$, the fraction of tasks for which at least one of $K$ sampled trajectories satisfies the TL specification. On Panda task, our model achieves $67.45\%$ Success@1024, compared with $59.11\%$ for the oracle-state baseline. On AntMaze task, it achieves $96.35\%$ Success@256, comparable to the oracle result of $96.88\%$. Resolution and intervention studies show that spatial detail depends on semantic grounding and predicate identity affects both attention and performance. These results demonstrate a direct mapping from visual observations and structured TL goals to action trajectories without requiring object geometry at inference. Code is available at \url{https://github.com/AricLau07/vision-tl-action}.
\end{abstract}

\section{Introduction}

Robotic tasks often couple goal achievement with deadlines, execution order, and safety constraints, requiring precise reasoning and planning to ensure correct, timely, and safe execution. 
For example, a robot may need to reach region $A$ within 10 seconds, avoid obstacle $B$ throughout the horizon $T$, and visit region $C$ only after reaching $A$. 
Signal temporal logic (STL) makes such requirements precise and machine-checkable over continuous trajectories~\citep{maler2004monitoring,belta2019formal}.

Although STL provides a rigorous criterion for trajectory validity, translating a specification into an executable trajectory remains a separate planning problem. 
Recent vision--language--action (VLA) policies learn to map visual observations and language instructions directly to robot actions~\citep{jiang2023vima,zitkovich2023rt2,kim2025openvla,black2024pi0}, but representative models do not explicitly condition action generation on compositional STL specifications.
A complementary line of research explores learning-based STL planning. These methods learn policies or generate trajectories from STL specifications~\citep{guo2024specification,meng2024diverse,liu2026zeroshotstl}, but generally rely on symbolic or state-based scene interfaces.
For example, TeLoGraF encodes general STL specifications as graphs to condition a flow-matching trajectory generator~\citep{meng2025telograf}. 
However, TeLoGraF relies on a privileged interface that stores exact metric geometry in its object-referenced nodes, thereby bypassing the predicate-to-scene grounding required under visual observation and analogous to phrase-to-region grounding in vision--language detection~\citep{liu2024grounding}.

To address the challenge of replacing exact object geometry with visual observations in STL-conditioned trajectory generation, we propose Vision--Temporal-Logic--Action (Vision-TL-Action).
The model encodes an STL graph without metric scene geometry into node tokens, while mapping visual observations to spatial visual tokens.
Bidirectional cross-attention enables mutual conditioning between the graph and visual tokens, while training-time grounding supervision encourages object-referenced predicates to attend to corresponding objects.
The resulting fused representation, together with the robot's initial state, conditions the flow-matching trajectory generator adopted from TeLoGraF.
Experiments on Panda and AntMaze show that visual conditioning can match or even surpass privileged conditioning in candidate-set coverage, despite weaker single-sample performance.

Our contributions are as follows:
(1) we introduce Vision-TL-Action, a visual extension of TeLoGraF that replaces exact object geometry with visual observations;
(2) we propose node-level graph--vision grounding using object-identity embeddings, bidirectional cross-attention, and training-time supervision; and
(3) we show on Panda and AntMaze that visual conditioning can match or surpass privileged conditioning, analyze grounding limitations, and validate our design through extensive ablations and attention visualizations.

\section{Related Work}
\paragraph{Temporal-logic planning and learning.}
Representative approaches to STL-based control synthesis and planning include
sampling-based synthesis~\citep{vasile2017sampling},
mixed-integer formulations and receding-horizon control~\citep{raman2015reactive,sadraddini2015robust},
and gradient-based optimization supported by differentiable implementations of
STL robustness~\citep{dawson2022robust,leung2023backpropagation,kapoor2025stlcg}.
Learning-based methods shift much of the computational cost of temporal-logic
synthesis to offline training. 
Representative approaches include neural predictive control for fixed STL requirements~\citep{meng2023signal},
specification-conditioned policies and planners~\citep{guo2024specification,feng2025diffusion,vaezipoor2021ltl2action},
and STL-parameter-conditioned diffusion~\citep{meng2024diverse}.
DeepLTL derives reach--avoid sequences from B\"uchi automata for unseen LTL
tasks~\citep{jackermeier2025deepltl}.
Other approaches guide diffusion samples at inference time~\citep{zhong2023guided,feng2024ltldog}
or compose trajectory segments from a pretrained model for unseen STL
formulas~\citep{liu2026zeroshotstl}.
TeLoGraF conditions flow matching on an STL syntax graph~\citep{meng2025telograf}; Vision-TL-Action removes exact object geometry from
this graph and replaces it with visual evidence.

\paragraph{Vision-conditioned trajectory generation.}
Generative planners model distributions over trajectories or actions, with
conditioning ranging from task variables to visual scene observations.
Diffuser plans by iteratively denoising state--action trajectories~\citep{janner2022planning},
whereas Diffusion Policy and FlowPolicy generate actions from RGB and 3-D
observations~\citep{chi2025diffusion,zhang2025flowpolicy}.
These methods do not explicitly bind STL predicates to image regions.
MDETR and GLIP provide analogous token--region grounding~\citep{Kamath_2021_ICCV,Li_2022_CVPR}, while S-MSP fuses multi-view RGB,
STL structure, and region information for waypoint generation but lacks
node-level grounding supervision~\citep{ye2026smsp}.
Vision-TL-Action explicitly aligns predicate nodes with visual tokens through
cross-attention and training-time supervision.

\paragraph{Vision--language--action models.}
VLA models predict actions from visual observations and instructions, including
robot transformers and VLM-based policies~\citep{brohan2023rt,zitkovich2023rt2,li2024roboflamingo},
cross-embodiment generalists~\citep{openx2024,ghosh2024octo,kim2025openvla},
and recent generative or simplified designs~\citep{black2024pi0,pi2025pi05,bjorck2025groot,ye2026starvlaalpha}.
Some further introduce 3-D or spatial representations~\citep{zhen2024threedvla,qu2025spatialvla}.
However, these models generally do not encode compositional STL syntax or
explicit predicate-to-region bindings.

\section{Problem Formulation}

Let 
$\boldsymbol{\tau}=\left(
\boldsymbol{q}_0 , \boldsymbol{a}_0, \boldsymbol{q}_1, \boldsymbol{a}_1, \ldots , \boldsymbol{q}_H \right)$ 
denote a trajectory, where
$\boldsymbol{q}_t\in\mathbb{R}^n$ and
$\boldsymbol{a}_t\in\mathbb{R}^m$ denote the system state and action at time $t$,
respectively, and $\mathcal{H}$ is the planning horizon. When an agent executes
$\boldsymbol{a}_t$ at $\boldsymbol{q}_t$, it transitions to $\boldsymbol{q}_{t+1}$.

\subsection{Signal Temporal Logic (STL)}
\label{subsec-stl}

Signal temporal logic (STL) specifies temporal constraints over a trajectory
$\boldsymbol{\tau}$ through atomic predicates (APs) and logical-temporal
operators. An atomic predicate $\mu$ is induced by a predicate function
$h_\mu$ defined over the state $\boldsymbol{q}_t$ and holds at time
$t$ if $h_\mu(\boldsymbol{q}_t)>0$. 
An STL formula $\Phi$ is defined recursively as follows~\citep{lindemann2021funnel}:
\begin{equation}
    \Phi ::= \top
    \mid \mu
    \mid \neg\Phi
    \mid \Phi_1\wedge\Phi_2
    \mid \Phi_1\,U_{[a,b]}\,\Phi_2,
    \label{eq:stl_syntax}
\end{equation}
where $\top$ denotes true, $\neg$ and $\wedge$ denote negation and
conjunction, and $U_{[a,b]}$ is the until operator over the interval $[a,b]$.
The eventually and always operators are derived as
$F_{[a,b]}\Phi=\top\,U_{[a,b]}\,\Phi$ and
$G_{[a,b]}\Phi=\neg F_{[a,b]}\neg\Phi$. We denote $\boldsymbol{\tau},t\models\Phi$ if the trajectory
$\boldsymbol{\tau}$ satisfies $\Phi$ at time $t$. For simplicity, define $\mathcal{I}_t := [t+a, t+b]$. The Boolean semantics are
defined recursively as
$\boldsymbol{\tau},t\models\top\Leftrightarrow\mathrm{True}$;
and $\boldsymbol{\tau},t\models\mu\Leftrightarrow
h_\mu(\boldsymbol{q}_t)>0$;
and $\boldsymbol{\tau},t\models\neg\Phi\Leftrightarrow
\boldsymbol{\tau},t\not\models\Phi$;
and $\boldsymbol{\tau},t\models\Phi_1\wedge\Phi_2\Leftrightarrow
(\boldsymbol{\tau},t\models\Phi_1)\wedge
(\boldsymbol{\tau},t\models\Phi_2)$;
and $\boldsymbol{\tau},t\models\Phi_1\,U_{[a,b]}\,\Phi_2$ iff there exists
$t'\in\mathcal{I}_t$ such that $\boldsymbol{\tau},t'\models\Phi_2$ and
$\boldsymbol{\tau},t''\models\Phi_1$ for all $t''\in[t,t']$;
and $\boldsymbol{\tau},t\models F_{[a,b]}\Phi$ iff there exists
$t'\in\mathcal{I}_t$ such that $\boldsymbol{\tau},t'\models\Phi$;
and $\boldsymbol{\tau},t\models G_{[a,b]}\Phi$ iff
$\boldsymbol{\tau},t'\models\Phi$ for all $t'\in\mathcal{I}_t$.
For an object-referenced predicate,
evaluating $h_\mu$ also requires the geometry of the corresponding target or
obstacle. 
For example, let $\boldsymbol{p}_t$ denote the robot position contained in
$\boldsymbol{q}_t$, and represent an object by its center
$\boldsymbol{o}$ and radius $r$. The STL formula
$F_{[t_1,t_2]}\mu_{\mathrm{reach}}$ requires the robot to reach
the object at some time between $t_1$ and $t_2$. The corresponding
predicate is defined as
$h_{\mathrm{reach}}(\boldsymbol{q}_t;\boldsymbol{o},r) = r^2-\left\|\boldsymbol{p}_t-\boldsymbol{o}\right\|_2^2$,
where $\mu_{\mathrm{reach}}$ is satisfied when
$h_{\mathrm{reach}}(\boldsymbol{q}_t;\boldsymbol{o},r)> 0$.

The robustness semantics $\rho^\Phi(\boldsymbol{\tau},t)$ quantify how well
$\boldsymbol{\tau}$ satisfies $\Phi$ at time $t$. We use
$\rho^\Phi(\boldsymbol{\tau},t)>0$ as the strict satisfaction criterion.
Let
$\rho_t^\Phi:=\rho^\Phi(\boldsymbol{\tau},t)$. The robustness semantics are defined recursively as follows~\citep{lindemann2019feedback}:
$\rho_t^\top=+\infty$;
and $\rho_t^\mu=h_\mu(\boldsymbol{q}_t)$;
and $\rho_t^{\neg\Phi}=-\rho_t^\Phi$;
and $\rho_t^{\Phi_1\wedge\Phi_2}
=\min\{\rho_t^{\Phi_1},\rho_t^{\Phi_2}\}$;
and $\rho_t^{F_{[a,b]}\Phi}
=\sup_{t'\in\mathcal{I}_t}\rho_{t'}^\Phi$;
and $\rho_t^{G_{[a,b]}\Phi}
=\inf_{t'\in\mathcal{I}_t}\rho_{t'}^\Phi$;
and $\rho_t^{\Phi_1\,U_{[a,b]}\,\Phi_2}
=\allowbreak\sup_{t'\in\mathcal{I}_t}
\min\{\rho_{t'}^{\Phi_2},
\allowbreak\inf_{t''\in[t,t']}\rho_{t''}^{\Phi_1}\}$.

\subsection{Vision-Conditioned Temporal-Logic Planning}
Following the syntax-graph representation of TeLoGraF~\citep{meng2025telograf},
we define a coordinate-free STL graph
\(\mathcal{G}_{\Phi}=(\mathcal{V}_{\Phi},\mathcal{E}_{\Phi},
\mathcal{F}_{\Phi})\), where \(\mathcal{V}_{\Phi}\) and
\(\mathcal{E}_{\Phi}\) are the node and directed syntax-edge sets, and
\(\mathcal{F}_{\Phi}=\{\boldsymbol{f}_v\}_{v\in\mathcal{V}_{\Phi}}\) is the
node-feature set. Each \(\boldsymbol{f}_v\) retains the operator type
\(L_{\mathrm{type}}\), temporal interval
\((t_{\mathrm{start}},t_{\mathrm{end}})\), atomic predicate type
\(L_{\mathrm{ap}}\), object identity \(L_{\mathrm{obj}}\), shape
category \(L_{\mathrm{shape}}\), and syntax-role indicator
\(L_{\mathrm{left}}\) for the ordered arguments of Until, but excludes exact
object coordinates, radii, and other metric scene geometry. 
Here, coordinate-free refers only to the task graph.
The environment is observed through \(C\) views
\(\mathcal{I}=\{I^{(c)}\}_{c=1}^{C}\), where each \(I^{(c)}\) is an RGB image
or a synthetic semantic raster. Given \(\mathcal{G}_{\Phi}\),
\(\mathcal{I}\), and the initial robot state \(\boldsymbol{q}_0\),
vision-conditioned temporal-logic planning generates trajectories
\(\boldsymbol{\tau}\) intended to satisfy \(\Phi\). The graph specifies the
referenced objects and predicate composition, the observations provide scene
evidence, and \(\boldsymbol{q}_0\) specifies the initial condition.

\subsection{Task Satisfaction and Evaluation}
We compute ground-truth STL robustness
$\rho^{\Phi}(\boldsymbol{\tau},0)$ offline using simulator-provided geometry.
For task $i$, given $K$ generated trajectories
$\{\boldsymbol{\tau}_{i,j}\}_{j=1}^{K}$, we report
\begin{align}
    S_i@K
    &=
    \mathds{1}\!\left[
        \max_{1\leq j\leq K}
        \rho^{\Phi_i}(\boldsymbol{\tau}_{i,j},0)>0
    \right],\\
    \mathrm{Success}@K
    &=
    \frac{1}{M}\sum_{i=1}^{M}S_i@K.
    \label{eq:success_at_k}
\end{align}
Here, $S_i@K$ indicates whether at least one of the $K$ candidates satisfies
$\Phi_i$, and $\mathrm{Success}@K$ averages this indicator over the $M$
evaluation tasks. 
Following prior STL-conditioned generative
planners~\citep{meng2025telograf,meng2024diverse,zhong2023guided}, 
we formulate $\mathrm{Success}@K$ as a best-of-$K$ evaluation metric for
stochastic conditional generation. Since satisfaction is determined offline
using ground-truth STL robustness, this metric quantifies the coverage of
satisfying trajectories rather than the performance of a deployable
trajectory-ranking mechanism.

\subsection{Problem Formulation}

We consider a dataset of demonstrated temporal-logic planning tasks
$\mathcal{D}=\left\{\left(\mathcal{I}_i,\Phi_i,
\{\boldsymbol{\tau}_{i,j}\}_{j=1}^{R}\right)\right\}_{i=1}^{W}$,
where $\mathcal{I}_i$$=\{I_i^{(c)}\}_{c=1}^{C}$ is the multi-view observation of task $i$, 
$\Phi_i$ is the STL formula defined in \eqref{eq:stl_syntax},
and $\{\boldsymbol{\tau}_{i,j}\}_{j=1}^{R}$ is a set of demonstrated trajectories satisfying the specification, i.e.,
$\boldsymbol{\tau}_{i,r},0\models\Phi_i$ for all $r$. 
We denote by $\mathcal{G}_{\Phi_i}$ the coordinate-free STL graph
constructed from $\Phi_i$. Our goal is to learn a conditional generative model
$p_\theta(\boldsymbol{\tau}\mid\boldsymbol{q}_0,\mathcal{I},
\mathcal{G}_{\Phi})$
such that, given a multi-view
observation $\mathcal{I}$, a coordinate-free STL graph
$\mathcal{G}_{\Phi}$, and an initial robot state $\boldsymbol{q}_0$, the model
samples $K$ candidate trajectories
$\boldsymbol{\tau}_j\sim
p_\theta(\cdot\mid\boldsymbol{q}_0,\mathcal{I},\mathcal{G}_{\Phi})$,
$j=1,\ldots,K$, with the goal that at least one candidate satisfies the
query STL specification $\Phi$, i.e.,
$\exists j\in\{1,\ldots,K\}$ such that
$\boldsymbol{\tau}_j,0\models\Phi$.

\section{Vision-TL-Action}
\begin{figure*}[t]
    \centering
    \includegraphics[width=0.9\textwidth]{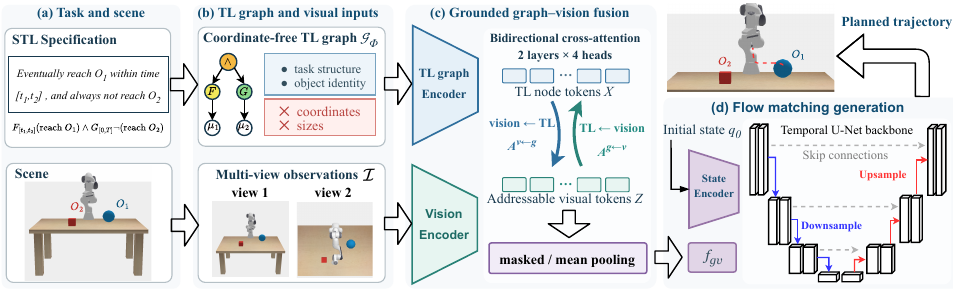}
    \caption{\textbf{System overview of Vision-TL-Action.}
    The model encodes a coordinate-free TL graph and multi-view visual
    observations into TL node and visual tokens, fuses them through
    bidirectional cross-attention, and conditions a
    flow-matching Temporal U-Net on the fused feature and encoded
    initial state to generate candidate trajectories by ODE integration.}
    \label{fig:example}
\end{figure*}

\subsection{Coordinate-Free TL Graph Encoding}
\label{subsec:coordinate_free_tl_graph_encoding}

Each object identity is first represented by a categorical index in the coordinate-free TL graph. We then map it to a learned continuous embedding before graph encoding. We define the object-identity embedding as
\begin{equation*}
    e_{\mathrm{obj}}:
    \mathbb{L}_{\mathrm{obj}}
    \rightarrow
    \mathbb{R}^{d_{\mathrm{obj}}},
    \qquad
    \mathbb{L}_{\mathrm{obj}}
    :=
    \{-1,0,\ldots,W-1\},
\end{equation*}
where $L_{\mathrm{obj}}\in\mathbb{L}_{\mathrm{obj}}$ is the object identity
associated with a TL node, $W$ is the maximum number of within-scene object
identities, and $d_{\mathrm{obj}}$ is the embedding dimension. We reserve
$L_{\mathrm{obj}}=-1$ for nodes that do not refer to an object. Replacing the
scalar object identity in $\boldsymbol{f}_v$ with
$e_{\mathrm{obj}}(L_{\mathrm{obj}})$ gives the initial node representation
\begin{equation*}
    \boldsymbol{f}_v^{(0)}
    =
    \left[
        L_{\mathrm{type}},
        t_{\mathrm{start}},
        t_{\mathrm{end}},
        L_{\mathrm{ap}},
        e_{\mathrm{obj}}(L_{\mathrm{obj}}),
        L_{\mathrm{shape}},
        L_{\mathrm{left}}
    \right].
\end{equation*}
The object-identity embedding distinguishes nodes that refer to different
scene objects without encoding their locations or sizes. The embedding parameters are learned during model training.

Following TeLoGraF~\citep{meng2025telograf}, we propagate information over
the child-to-parent syntax graph using an $L_{\mathrm{g}}$-layer
graph-convolutional encoder $E_{\mathrm{TL}}$, with ReLU activations after
all but the final layer. Rather than immediately aggregating the final node
representations into a single graph-level embedding, as in TeLoGraF, we
retain the final representation of each TL node as a separate token:
\begin{equation}
    X
    =
    E_{\mathrm{TL}}(\mathcal{G}_{\Phi})
    =
    [x_1,\ldots,x_N]^{\top}
    \in\mathbb{R}^{N\times d},
    \label{eq:coordinate_free_tl_tokens}
\end{equation}
where
$x_i=\boldsymbol{f}_{v_i}^{(L_{\mathrm{g}})}\in\mathbb{R}^{d}$
is the TL node token associated with node $v_i$,
$N=|\mathcal{V}_{\Phi}|$ is the number of TL nodes, and $d$ is the token
dimension. The resulting TL node-token sequence $X$ is passed to
graph--vision cross-attention, allowing each object-referenced AP node to
query scene-dependent spatial evidence from the visual observations before
pooling.

\subsection{Addressable Multi-View Visual Tokens}
\label{subsec:addressable_visual_tokens}

We encode the multi-view observation
$\mathcal{I}=\{I^{(c)}\}_{c=1}^{C}$ into local visual tokens while
preserving their feature-grid locations and source-view information.
A shared visual backbone processes each view independently. Let
$
B_{\mathrm{vis}}:
\mathbb{R}^{3\times \mathcal{H}\times \mathcal{W}}
\rightarrow
\mathbb{R}^{d_f\times h\times w}
$
denote this backbone. For view $c$ and feature-grid index
$p\in\{1,\ldots,hw\}$, the corresponding local feature is
\begin{equation*}
    f_{c,p}
    =
    \left[
        \operatorname{Flat}_{hw}
        \!\left(B_{\mathrm{vis}}(I^{(c)})\right)
    \right]_p
    \in\mathbb{R}^{d_f},
\end{equation*}
where $\operatorname{Flat}_{hw}$ converts the $h\times w$ feature grid into
an ordered sequence. We concatenate the features in view-major order and
define $f_j=f_{c,p}$, $j=(c-1)hw+p$, $L=Chw$.
A shared linear projection maps each local feature to the token dimension
$d$ used by the TL node tokens:
\[
    e_j=W_e f_j+b_e\in\mathbb{R}^{d},
    \qquad
    E=[e_1,\ldots,e_L]^\top\in\mathbb{R}^{L\times d}.
\]

To preserve the grid location and source view after flattening, we assign
each token an address $s_j=[\xi_j,\eta_j,\nu_j]^\top$,
where $(\xi_j,\eta_j)\in[-1,1]^2$ denotes its normalized feature-grid
coordinates and $\nu_j\in[-1,1]$ denotes its normalized view coordinate.
We project this address to the shared token dimension and add it to the
corresponding content token:
\begin{equation}
    z_j=e_j+W_s s_j+b_s,
    \qquad
    Z=[z_1,\ldots,z_L]^\top\in\mathbb{R}^{L\times d}.
    \label{eq:spatial_token}
\end{equation}
We initialize $W_s$ and $b_s$ to zero and learn them during model training.
The resulting sequence $Z$ is spatially addressable through normalized grid and view coordinates; the address encoding contains neither symbolic object identities nor metric world coordinates.

\subsection{Graph--Vision Cross-Attention}
\label{subsec:graph_visual_cross_attention}

Separate encodings of the TL graph and visual observations do not identify
which visual evidence corresponds to each object-referenced predicate.
Before pooling, we fuse the TL node tokens $X$ and addressable
visual tokens $Z$ through node-level bidirectional graph--vision
cross-attention.

Let $X^{(0)}=X$ and $Z^{(0)}=Z$, and let $\operatorname{LN}$ and
$\operatorname{MHA}(Q,K,V)$ denote layer normalization and multi-head
attention, respectively. The module has two bidirectional layers
$\ell\in\{1,2\}$. In each direction, it uses four heads
$h\in\{1,\ldots,4\}$ with dimension $d_h=d/4$.

Each layer first updates the TL node-token matrix:
\begin{equation*}
\begin{aligned}
X^{(\ell)}
&=
\operatorname{LN}\!\Bigl(
    X^{(\ell-1)}
    {}\\[-2pt]
&\qquad
    {}+\operatorname{MHA}_{\ell}^{g\leftarrow v}
    \bigl(
        X^{(\ell-1)},
        Z^{(\ell-1)},
        Z^{(\ell-1)}
    \bigr)
\Bigr).
\end{aligned}
\end{equation*}
Each TL node token thus retrieves visual evidence through a residual update.
For head $h$, the TL-to-vision attention matrix
$A_{\ell,h}^{g\leftarrow v}\in\mathbb{R}^{N\times L}$ is
\begin{equation*}
\begin{aligned}
A_{\ell,h}^{g\leftarrow v}
=
\operatorname{softmax}\!\left(
\frac{
\bigl(X^{(\ell-1)}W_{Q,\ell,h}^{g}\bigr)
\bigl(Z^{(\ell-1)}W_{K,\ell,h}^{v}\bigr)^\top
}{
\sqrt{d_h}
}
\right).
\end{aligned}
\end{equation*}
Each row gives the attention distribution from one TL node token to the
$L$ visual tokens.
The updated TL node tokens then condition the visual-token matrix:
\begin{equation*}
\begin{aligned}
Z^{(\ell)}
&=
\operatorname{LN}\!\Bigl(
    Z^{(\ell-1)}
    +\operatorname{MHA}_{\ell}^{v\leftarrow g}
    \bigl(
        Z^{(\ell-1)},
        X^{(\ell)},
        X^{(\ell)}
    \bigr)
\Bigr).
\end{aligned}
\end{equation*}
In this reverse direction,
$A_{\ell,h}^{v\leftarrow g}\in\mathbb{R}^{L\times N}$
denotes the head-$h$ vision-to-TL attention matrix, whose rows give
distributions from visual tokens to the $N$ updated TL node tokens.

The pair $(X^{(\ell)},Z^{(\ell)})$ enters the next layer. After two layers,
the module outputs $\widetilde{X}=X^{(2)}$ and
$\widetilde{Z}=Z^{(2)}$. We retain the final-layer TL-to-vision attention
$A_{2,h}^{g\leftarrow v}$ for the grounding objective defined next.
Finally, we pool the updated token sequences into the fused graph--vision
feature
\begin{equation}
f_{gv}
=
W_o\!\left[
    \operatorname{Pool}_{\mathrm{valid}}(\widetilde{X})
    +
    \operatorname{Mean}(\widetilde{Z})
\right],
\label{eq:graph_visual_feature}
\end{equation}
where $\operatorname{Pool}_{\mathrm{valid}}$ is a masked mean over valid TL
node tokens, $\operatorname{Mean}$ averages the visual tokens, and $W_o$ is
a learned post-pooling projection.

\subsection{Predicate-to-Object Grounding}
\label{subsec:predicate_to_object_grounding}

Cross-attention allows object-referenced AP node tokens to attend to visual
tokens, but trajectory-level supervision does not directly specify their
correspondence. We therefore introduce an auxiliary predicate-to-object
grounding objective that directly supervises attention to the referenced
object locations.
For each object-referenced AP node $i$, let
$k(i)\in\mathbb{L}_{\mathrm{obj}}\setminus\{-1\}$ denote its object identity,
and let $\mathcal{Q}$ contain the AP nodes with at least one valid grounding
target. During training, an environment-specific geometric mapping assigns
each referenced object center to its visual-token indices in the designated
supervised views. The resulting valid indices define
$\mathcal{R}_{k(i)}\subseteq\{1,\ldots,L\}$.
We first average the final-layer TL-to-vision attention over its four heads:
\begin{equation}
\overline{A}_{ij}
=
\frac{1}{4}
\sum_{h=1}^{4}
\left[A_{2,h}^{g\leftarrow v}\right]_{ij}.
\end{equation}
The grounding loss is then
\begin{equation}
\mathcal{L}_{\mathrm{ground}}
=
-\frac{1}{|\mathcal{Q}|}
\sum_{i\in\mathcal{Q}}
\log\!\left(
\max\!\left\{
\sum_{j\in\mathcal{R}_{k(i)}}
\overline{A}_{ij},
\epsilon
\right\}
\right),
\label{eq:grounding_loss}
\end{equation}
where $\epsilon=10^{-8}$ lower-bounds the target attention mass for
numerical stability. If $\mathcal{Q}$ is empty, we set
$\mathcal{L}_{\mathrm{ground}}=0$. This objective increases the total
attention mass assigned to the projected object-center tokens across the
supervised views. The projections and target-token indices provide
privileged training-only supervision and are not used at inference.

\subsection{Flow-Matching Trajectory Generation}

Following TeLoGraF~\citep{meng2025telograf}, we use its conditional
flow-matching Temporal U-Net for trajectory generation. Let $e_0=E_q(\boldsymbol{q}_0)\in\mathbb{R}^{d_q}$, $c=[f_{gv};e_0]\in\mathbb{R}^{d+d_q}$
denote the initial-state embedding and the complete conditioning vector,
respectively. Let $\mathbb{T}$ denote the original trajectory space and
$\widetilde{\mathbb{T}}$ its normalized representation. We learn the
conditional vector field
$v_{\theta}:
\mathbb{R}^{d+d_q}\times[0,1]\times\widetilde{\mathbb{T}}
\rightarrow
\widetilde{\mathbb{T}},$
whose induced flow produces a normalized trajectory that is mapped back to
a candidate $\widehat{\boldsymbol{\tau}}\in\mathbb{T}$ intended to satisfy
$\Phi$.

At each training step, we sample a demonstration
$\boldsymbol{\tau}_1$ from the training data and normalize it as
$\widetilde{\boldsymbol{\tau}}_1
=\operatorname{Norm}(\boldsymbol{\tau}_1)$. We also sample a Gaussian source
trajectory
$\boldsymbol{\tau}_0\sim\mathcal{N}(0,I)$ and a generative flow time
$s\sim\operatorname{Uniform}(0,1)$. The interpolated trajectory and target
velocity are
\begin{equation}
\widetilde{\boldsymbol{\tau}}_s
=
s\widetilde{\boldsymbol{\tau}}_1
+
(1-s)\boldsymbol{\tau}_0,
\qquad
\Delta\boldsymbol{\tau}
=
\widetilde{\boldsymbol{\tau}}_1-\boldsymbol{\tau}_0.
\end{equation}
The flow-matching objective is
\begin{equation}
\mathcal{L}_{\mathrm{FM}}
=
\mathbb{E}_{\boldsymbol{\tau}_1,\boldsymbol{\tau}_0,s}
\left[
\left\|
v_{\theta}(c,s,\widetilde{\boldsymbol{\tau}}_s)
-\Delta\boldsymbol{\tau}
\right\|_2^2
\right].
\label{eq:flow_matching_loss}
\end{equation}
Together with the predicate-to-object grounding objective in
Equation~\ref{eq:grounding_loss}, the overall training objective is
\begin{equation}
\mathcal{L}
=
\mathcal{L}_{\mathrm{FM}}
+
\lambda_{\mathrm{g}}\mathcal{L}_{\mathrm{ground}},
\label{eq:overall_training_loss}
\end{equation}
where $\lambda_{\mathrm{g}}$ weights the auxiliary grounding objective.

At inference, we construct $c$ from the initial state
$\boldsymbol{q}_0$, visual observation $\mathcal{I}$, and coordinate-free
TL graph $\mathcal{G}_{\Phi}$. Starting from
$\boldsymbol{\psi}_0\sim\mathcal{N}(0,I)$, we evolve the normalized
trajectory according to
$\frac{\mathrm{d}\boldsymbol{\psi}_s}{\mathrm{d}s}
=
v_{\theta}(c,s,\boldsymbol{\psi}_s)$, $s\in[0,1]$.
We solve this ODE with $N_s$ Euler steps. Here, $s$ denotes generative time
from the Gaussian source to the data distribution; the implementation uses
the equivalent reverse-indexed discrete time convention of the inherited
TeLoGraF backbone. The final candidate is
$\widehat{\boldsymbol{\tau}}
=
\operatorname{Denorm}(\boldsymbol{\psi}_{N_s})$,
where $\operatorname{Denorm}$ inverts the training-time normalization and
returns the sample to the original trajectory coordinates.

\section{Experiments}
Our experiments evaluate five aspects of Vision-TL-Action: (1) the ability of
visual observations to replace exact object geometry while preserving
TL-conditioned trajectory coverage; (2) the transition from single-sample
generation to oracle-evaluated candidate-set coverage as the number of generated
trajectories increases; (3) the effect of node-level grounding on planning and
predicate--object correspondence; (4) the extent to which the final generators
rely on visual evidence rather than nonvisual task priors; and (5) whether
cross-attention follows TL object identity under graph-only interventions in
both domains.

\subsection{Experimental Setup}

\paragraph{Domains and protocol.}
We evaluate Vision-TL-Action in the complementary simulated domains of Panda
manipulation and AntMaze navigation. In the Panda manipulation, the
filtered dataset contains 71,325 trajectory samples that satisfy
object-referenced TL specifications involving reaching, avoidance, ordering,
and temporal composition in tabletop scenes. These samples are split into
57,058 training and 14,267 validation samples. The source image index contains
225,860 rendered views from the custom and bird's-eye cameras. Evaluation uses
a fixed offset-128 holdout of 128 unique TL tasks at candidate set sizes
$K\in\{1,16,64,128,256,512,1024\}$.
In the AntMaze navigation domain, a planar agent generates trajectories that
satisfy TL specifications involving reachability, avoidance, sequential goals,
and until constraints in a fixed-wall maze. The dataset contains 39,805
training and 10,071 validation trajectory samples. Visual input comprises two
$64\times64$ semantic rasters, one encoding the maze map and the other the task
targets; each is represented as a $16\times16$ token grid. The model retains
the 29-dimensional initial observation as nonvisual conditioning. Evaluation
uses 64 fixed unique TL tasks at candidate set sizes
$K\in\{1,16,64,128,256\}$.

\paragraph{Controls and statistics.}
The oracle-state control receives exact object geometry, while every visual
model receives the coordinate-free TL graph. For a fair comparison, all
methods are evaluated on identical task orderings using sampling seeds 1007,
2027, and 3407. For each task and seed, $\mathrm{Success}@K$ is computed from
the first $K$ trajectories of a fixed candidate sequence, yielding nested
candidate sets across sampling budgets. We report the mean and sample
standard deviation across seeds. Pairwise effects are computed after averaging
seeds within each task and bootstrapping tasks 20,000 times. Attention effects
use task-paired bootstrap intervals and two-sided sign-flip tests.

\paragraph{Implementation details.}
Both domains use node-level TL tokens, object-ID embeddings, two fusion layers,
four attention heads, and the same flow-matching Temporal U-Net. We use
$\lambda_{\mathrm{g}}=0.003$ for Panda and $0.002$ for AntMaze. Panda uses
batch size 256 and a five-epoch grounding stage. AntMaze uses batch size 128:
the native $16\times16$ model is first trained for 100 epochs (31,100 updates),
then trained for 30 additional epochs (9,330 updates) while unfreezing only
the existing condition/time projections in the trajectory generator. This
limited adaptation rule is shared with Panda; no domain-specific generator is
introduced.

\begin{table*}[t]
\centering
\small
\begin{tabular}{llccc}
\toprule
Domain & Condition & Object Coordinates at Test & Set Size $K$ & Success@$K$ (\%) \\
\midrule
\multirow{3}{*}{Panda}
& Oracle state & Yes & 1024 & $59.11\pm1.19$ \\
& Vision-TL fusion & No & 1024 & $62.76\pm4.58$ \\
& Vision-TL-Action (ours) & No & 1024 & $\mathbf{67.45\pm1.63}$ \\
\midrule
\multirow{2}{*}{AntMaze}
& Oracle state & Yes & 256 & $96.88\pm1.56$ \\
& Vision-TL-Action, $16\times16$ (ours) & No & 256 & $\mathbf{96.35\pm0.90}$ \\
\bottomrule
\end{tabular}
\caption{Three-seed trajectory coverage on fixed task sets. The proposed
coordinate-free model improves over the visual fusion model on Panda and
closes the gap to exact object state on AntMaze.}
\label{tab:main_results}
\end{table*}

\subsection{Performance--Coverage Analysis}
Table~\ref{tab:main_results} summarizes the main planning results.
On Panda,
Vision-TL-Action reaches $67.45\%\pm1.63$ Success@1024. The paired improvement
over the visual fusion model is $+4.69$ percentage points (pp), with a 95\%
bootstrap interval of $[1.56,8.07]$ pp. Relative to oracle state, the
improvement is $+8.33$ pp, with interval $[1.56,15.36]$ pp. Thus, object
geometry is not required to obtain competitive conditional generation in the
manipulation domain.

On AntMaze, the native $16\times16$ model reaches
$96.35\%\pm0.90$ Success@256, compared with $96.88\%\pm1.56$ for oracle state.
The paired difference is $-0.52$ pp with a 95\% bootstrap interval of
$[-4.69,3.65]$ pp. The interval crosses zero, so we claim that the proposed
model closes the observed state gap, not that it significantly outperforms the
oracle-state control. Appendix~\ref{sec:resolution_diagnostics} separately
records the representation-resolution studies that led to this final design.

\begin{figure}[!ht]
\centering
\includegraphics[width=0.9\columnwidth]{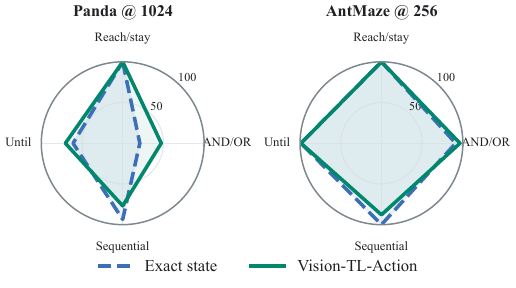}
\caption{Three-seed mean Success@$K$ by TL formula family at the maximum
candidate set size. The four axes are reach/stay, conjunction/disjunction,
sequential, and until tasks; values are percentages on the same fixed task
partitions used in Table~\ref{tab:main_results}.}
\label{fig:task_family_radar}
\end{figure}

Figure~\ref{fig:task_family_radar} resolves the aggregate result by formula
family. Vision-TL-Action improves conjunction/disjunction and until tasks on
Panda and conjunction/disjunction tasks on AntMaze, while exact state retains a
sequential-task advantage in both domains. Detailed task counts and uncertainty
appear in Appendix~\ref{sec:task_robustness}.

\begin{figure}[t]
\vspace{-5pt}
\centering
\includegraphics[width=0.75\columnwidth]{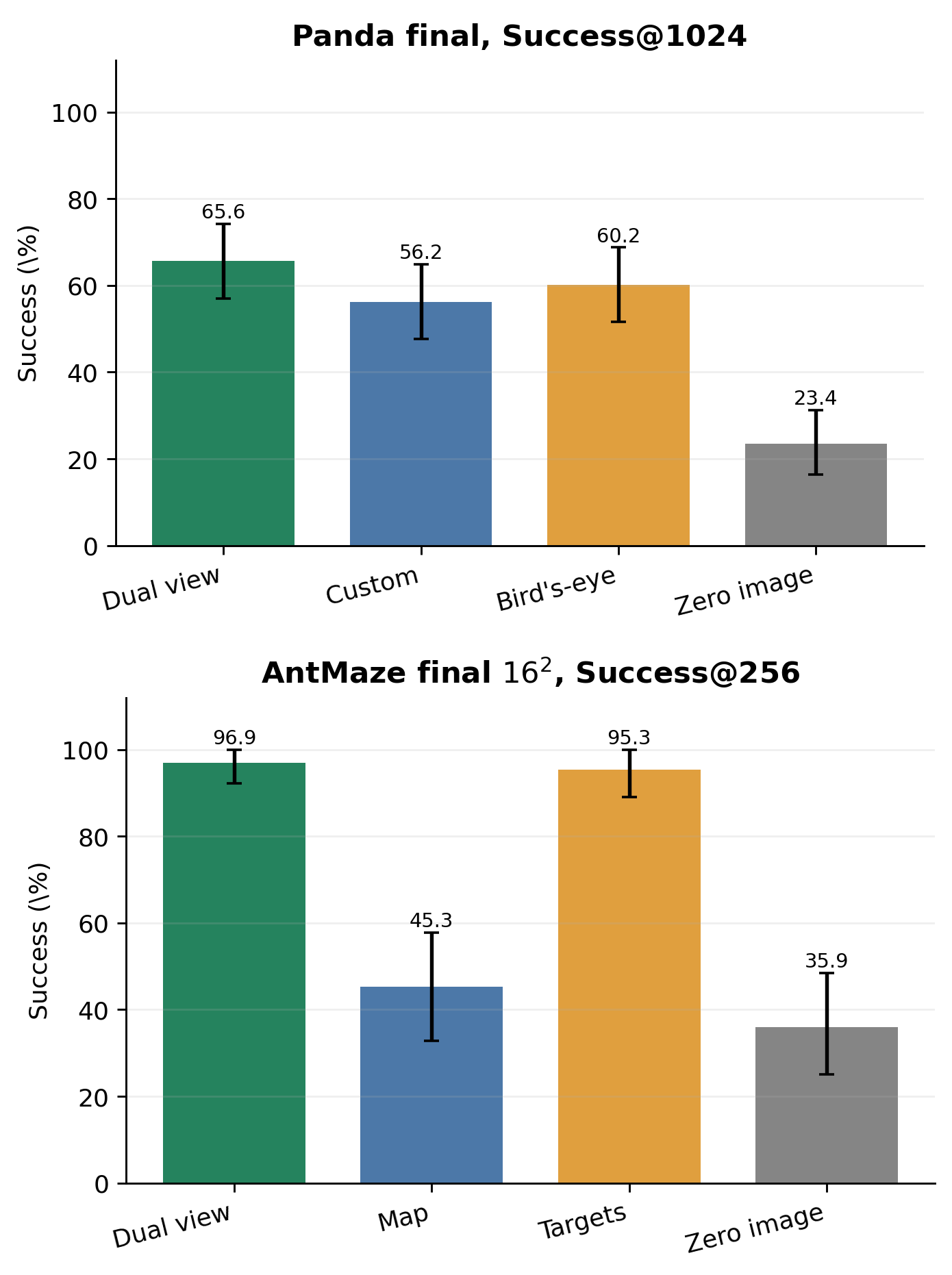}
\caption{Final-checkpoint visual-input interventions at seed 1007. Bars report
fixed-task success and error bars are 95\% task-bootstrap intervals. Panda
retains substantial information in either camera, whereas AntMaze depends
primarily on the task-specific target raster.}
\label{fig:visual_inputs}
\vspace{-5pt}
\end{figure}

\subsection{Visual-Input Intervention}
Figure~\ref{fig:visual_inputs} summarizes the fixed-checkpoint
visual-input interventions. We change only the visual inputs without
retraining, while holding the formal task, initial state, task order,
candidate set size, and sampling seed 1007 fixed. 
On Panda, dual-view Success@1024 is $65.62\%$, compared with $56.25\%$ for the
custom camera, $60.16\%$ for bird's-eye view, and $23.44\%$ for zero images.
Relative to dual view, the paired drops are $9.38$ pp ($[2.34,16.41]$),
$5.47$ pp ($[-0.78,11.72]$), and $42.19$ pp ($[33.59,50.78]$), respectively.
The bird's-eye camera carries most of the single-view signal, while the custom
camera contributes complementary evidence.

On AntMaze, dual-view Success@256 is $96.88\%$. Target-only input retains $95.31\%$, with a paired difference of $1.56$ pp ($[-3.12,6.25]$), whereas map-only and zero-image controls fall to $45.31\%$ and $35.94\%$. The targets-minus-map effect is $50.00$ pp ($[37.50,62.50]$). This weak marginal dependence on the map view is expected because the maze layout is fixed and its structural regularities are already implicit in the training distribution, allowing the trajectory generator to absorb much of the map information into its learned prior. It should therefore not be interpreted as evidence that map geometry is intrinsically irrelevant. In contrast, target configurations vary across tasks and provide the essential generalization condition that determines which behavior should be generated. The intervention results confirm this distinction: task-specific target appearance supplies the decisive visual semantics, whereas the fixed map primarily provides structural context. All intervals are 95\% task-bootstrap intervals.

\begin{figure}[!ht]
\centering
\includegraphics[width=0.8\columnwidth]{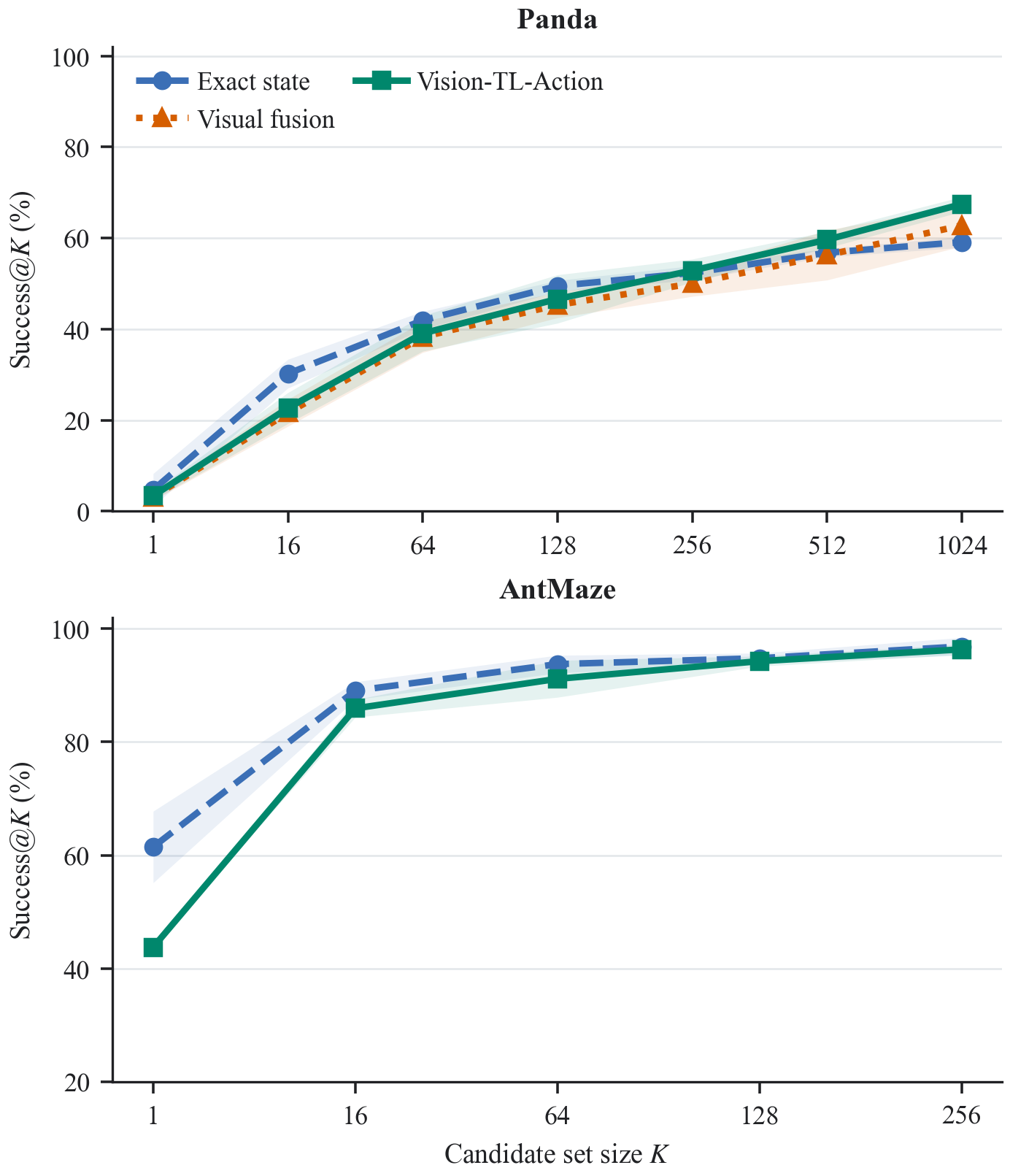}
\caption{Three-seed Success@$K$ over shared candidate prefixes. Exact state
produces stronger single samples, especially on AntMaze, while the proposed
visual model closes or reverses the gap as candidate diversity increases. The
bands show one sample standard deviation across sampling seeds. Panda also
includes its same-resolution visual-fusion baseline.}
\label{fig:candidate_budgets}
\end{figure}

\subsection{Single Samples versus Candidate Coverage}
Figure~\ref{fig:candidate_budgets} separates two properties that a single
maximum-set-size number conflates. On Panda, Success@1 is $3.39\%$ for our model
and $4.69\%$ for exact state; at $K=1024$, these values become $67.45\%$ and
$59.11\%$. On AntMaze, the single-sample gap is larger
($43.75\%$ versus $61.46\%$), but narrows to $96.35\%$ versus $96.88\%$ by
$K=256$. The proposed model therefore learns a competitive support over
satisfying trajectories, but its samples are less concentrated on success than
those of the exact-state model. This finding motivates a learned visual
trajectory ranker; it also prevents interpreting maximum-set coverage as
deployable top-1 performance.

\begin{figure*}[t]
\centering
\includegraphics[width=0.96\textwidth]{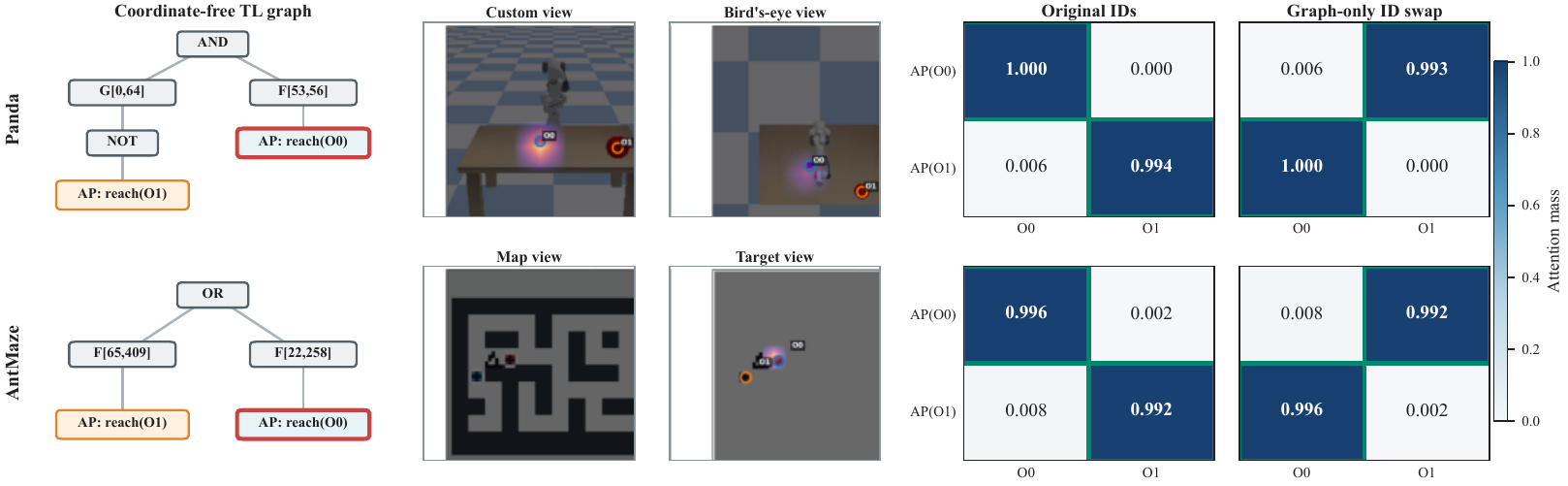}
\caption{Graph-to-vision grounding in Panda (top) and AntMaze (bottom). The TL
nodes state their operator, interval, predicate, and object identity. The two
visual columns show Panda custom and bird's-eye views and AntMaze map and target
views.
Outlined matrix cells mark the graph-consistent object; a graph-only identity
exchange moves the dominant correspondence while both visual views remain fixed.}
\label{fig:grounding_compact}
\end{figure*}

\subsection{TL Graph Grounding in Vision}
We audit all tasks, both fusion layers, and all four heads. Three metrics avoid
relying on a single heat map. \emph{Correct-object dominance} measures how
often the referenced object receives more attention than competing objects.
\emph{Top-5 hit} checks whether its region contains one of the five
highest-attended tokens. \emph{ID-swap shift} exchanges two object identities
in the TL graph while keeping pixels fixed and measures whether attention
preference follows the new identity.

On Panda, grounding increases dominance from 0.231 to 0.341, top-5 hit from
0.487 to 0.978, and ID-swap shift from 0.019 to 0.317. Task-paired improvements
are significant for dominance ($p=.0061$), top-5 hit ($p<10^{-4}$), and
ID-swap shift ($p<10^{-4}$). The final AntMaze model reaches 0.815 dominance,
0.999 top-5 hit, and a 1.164 ID-swap shift. These full-task statistics show
that localized predicate-object correspondence is not confined to selected
visual examples.

Figure~\ref{fig:grounding_compact} exposes the intervention itself. Each row
aligns a coordinate-free TL query, both visual views, and AP-to-object attention
mass before and after an object-ID exchange. With pixels fixed, the dominant
mass moves to the newly referenced object. The shown Panda and AntMaze
cases have cross-object Jensen--Shannon separations of 0.976 and 0.958 and
ID-swap preference shifts of 1.987 and 1.979, respectively. Full dual-view
panels, task-paired intervals, and retained failures appear in
Appendix~\ref{sec:extended_grounding}.

\subsection{Scope of the Current Evaluation}
The experiments establish coordinate-free visual conditioning at inference, but they do not remove all privileged information from training or evaluation. Object projections supervise Equation~\ref{eq:grounding_loss} during training, and Success@$K$ uses the ground-truth TL robustness to test candidate-set
coverage. AntMaze additionally uses synthetic semantic rasters rather than unannotated camera images. The current claim is therefore an end-to-end
visual-TL \emph{trajectory generator}; a fully deployable system additionally requires a learned candidate ranker, closed-loop replanning, and validation under real visual noise.

\section{Conclusion and Limitations}
Vision-TL-Action replaces metric object state in a TL graph with visual scene evidence while retaining compositional temporal structure. Across two domains, the model approaches or exceeds the oracle-state baseline for large candidate sets, although exact state remains stronger for individual samples. Visual ablations and graph-only identity interventions show that the generator uses task-relevant pixels and that predicate identity affects object-level
attention. These findings support visual--symbolic grounding.

Future work should extend the one-shot generator into a closed-loop system that repeatedly observes, reasons, acts, and replans, and validate it on physical robots under realistic visual and dynamics variations. Both directions require scalable robot data collection and simulation-to-real transfer, as large-scale datasets pairing visual observations, temporal tasks, actions, and outcomes remain scarce.

\FloatBarrier
\clearpage
\bibliography{aaai2027}

@inproceedings{meng2025telograf,
  title     = {{TeLoGraF}: Temporal Logic Planning via Graph-encoded Flow Matching},
  author    = {Meng, Yue and Fan, Chuchu},
  booktitle = {Proceedings of the 42nd International Conference on Machine Learning},
  series    = {Proceedings of Machine Learning Research},
  volume    = {267},
  pages     = {43754--43780},
  year      = {2025},
  publisher = {PMLR},
}

@inproceedings{maler2004monitoring,
  title     = {Monitoring Temporal Properties of Continuous Signals},
  author    = {Maler, Oded and Nickovic, Dejan},
  booktitle = {Formal Techniques, Modelling and Analysis of Timed and Fault-Tolerant Systems},
  series    = {Lecture Notes in Computer Science},
  volume    = {3253},
  pages     = {152--166},
  year      = {2004},
  publisher = {Springer},
  doi       = {10.1007/978-3-540-30206-3_12}
}

@inproceedings{zitkovich2023rt2,
  title={Rt-2: Vision-language-action models transfer web knowledge to robotic control},
  author={Zitkovich, Brianna and Yu, Tianhe and Xu, Sichun and Xu, Peng and Xiao, Ted and Xia, Fei and Wu, Jialin and Wohlhart, Paul and Welker, Stefan and Wahid, Ayzaan and others},
  booktitle={Conference on Robot Learning},
  pages={2165--2183},
  year={2023},
  organization={PMLR}
}

@inproceedings{kim2025openvla,
  title={OpenVLA: An Open-Source Vision-Language-Action Model},
  author={Kim, Moo Jin and Pertsch, Karl and Karamcheti, Siddharth and Xiao, Ted and Balakrishna, Ashwin and Nair, Suraj and Rafailov, Rafael and Foster, Ethan P and Sanketi, Pannag R and Vuong, Quan and others},
  booktitle={Conference on Robot Learning},
  pages={2679--2713},
  year={2025},
  organization={PMLR}
}

@article{black2024pi0,
  title={$\pi_0$: A Vision-Language-Action Flow Model for General Robot Control},
  author={Black, Kevin and Brown, Noah and Driess, Danny and Esmail, Adnan and Equi, Michael and Finn, Chelsea and Fusai, Niccolo and Groom, Lachy and Hausman, Karol and Ichter, Brian and others},
  journal={arXiv preprint arXiv:2410.24164},
  year={2024}
}

@inproceedings{raman2015reactive,
  title={Reactive synthesis from signal temporal logic specifications},
  author={Raman, Vasumathi and Donz{\'e}, Alexandre and Sadigh, Dorsa and Murray, Richard M and Seshia, Sanjit A},
  booktitle={Proceedings of the 18th international conference on hybrid systems: Computation and control},
  pages={239--248},
  year={2015}
}

@article{leung2023backpropagation,
  title={Backpropagation through signal temporal logic specifications: Infusing logical structure into gradient-based methods},
  author={Leung, Karen and Ar{\'e}chiga, Nikos and Pavone, Marco},
  journal={The International Journal of Robotics Research},
  volume={42},
  number={6},
  pages={356--370},
  year={2023},
  publisher={SAGE Publications Sage UK: London, England}
}

@inproceedings{guo2024specification,
author = {Guo, Zijian and Zhou, Weichao and Li, Wenchao},
title = {Temporal logic specification-conditioned decision transformer for offline safe reinforcement learning},
year = {2024},
publisher = {JMLR.org},
booktitle = {Proceedings of the 41st International Conference on Machine Learning},
articleno = {676},
numpages = {17},
location = {Vienna, Austria},
series = {ICML'24}
}

@article{meng2024diverse,
  title={Diverse Controllable Diffusion Policy with Signal Temporal Logic},
  author={Meng, Yue and Fan, Chuchu},
  journal={IEEE Robotics and Automation Letters},
  year={2024},
  publisher={IEEE}
}

@article{kapoor2025stlcg,
  title={Stlcg++: A masking approach for differentiable signal temporal logic specification},
  author={Kapoor, Parv and Mizuta, Kazuki and Kang, Eunsuk and Leung, Karen},
  journal={IEEE Robotics and Automation Letters},
  year={2025},
  publisher={IEEE}
}

@article{feng2024ltldog,
  title   = {{LTLDoG}: Satisfying Temporally-Extended Symbolic Constraints for Safe Diffusion-Based Planning},
  author  = {Feng, Zeyu and Luan, Hao and Goyal, Pranav and Soh, Harold},
  journal = {IEEE Robotics and Automation Letters},
  volume  = {9},
  number  = {10},
  pages   = {8571--8578},
  year    = {2024},
  doi     = {10.1109/LRA.2024.3443501}
}

@article{liu2026zeroshotstl,
  title={Zero-shot trajectory planning for signal temporal logic tasks},
  author={Liu, Ruijia and Hou, Ancheng and Yu, Xiao and Yin, Xiang},
  journal={Advances in Neural Information Processing Systems},
  volume={38},
  pages={130405--130442},
  year={2026}
}

@article{chi2025diffusion,
  title={Diffusion policy: Visuomotor policy learning via action diffusion},
  author={Chi, Cheng and Xu, Zhenjia and Feng, Siyuan and Cousineau, Eric and Du, Yilun and Burchfiel, Benjamin and Tedrake, Russ and Song, Shuran},
  journal={The International Journal of Robotics Research},
  volume={44},
  number={10-11},
  pages={1684--1704},
  year={2025},
  publisher={Sage Publications Sage UK: London, England}
}

@inproceedings{zhang2025flowpolicy,
  title={Flowpolicy: Enabling fast and robust 3d flow-based policy via consistency flow matching for robot manipulation},
  author={Zhang, Qinglun and Liu, Zhen and Fan, Haoqiang and Liu, Guanghui and Zeng, Bing and Liu, Shuaicheng},
  booktitle={Proceedings of the AAAI Conference on Artificial Intelligence},
  volume={39},
  number={14},
  pages={14754--14762},
  year={2025}
}

@inproceedings{Kamath_2021_ICCV,
  title={Mdetr-modulated detection for end-to-end multi-modal understanding},
  author={Kamath, Aishwarya and Singh, Mannat and LeCun, Yann and Synnaeve, Gabriel and Misra, Ishan and Carion, Nicolas},
  booktitle={Proceedings of the IEEE/CVF international conference on computer vision},
  pages={1780--1790},
  year={2021}
}

@inproceedings{Li_2022_CVPR,
  title={Grounded language-image pre-training},
  author={Li, Liunian Harold and Zhang, Pengchuan and Zhang, Haotian and Yang, Jianwei and Li, Chunyuan and Zhong, Yiwu and Wang, Lijuan and Yuan, Lu and Zhang, Lei and Hwang, Jenq-Neng and others},
  booktitle={Proceedings of the IEEE/CVF conference on computer vision and pattern recognition},
  pages={10965--10975},
  year={2022}
}

@misc{ye2026smsp,
  title={Bridging Perception and Planning: Towards End-to-End Planning for Signal Temporal Logic Tasks},
  author={Ye, Bowen and Huang, Junyue and Liu, Yang and Qiao, Xiaozhen and Yin, Xiang},
  journal={arXiv preprint arXiv:2509.12813},
  year={2025}
}

@article{belta2019formal,
  title   = {Formal Methods for Control Synthesis: An Optimization Perspective},
  author  = {Belta, Calin and Sadraddini, Sadra},
  journal = {Annual Review of Control, Robotics, and Autonomous Systems},
  volume  = {2},
  number  = {1},
  pages   = {115--140},
  year    = {2019},
  doi     = {10.1146/annurev-control-053018-023717}
}

@inproceedings{vasile2017sampling,
  title={Sampling-based synthesis of maximally-satisfying controllers for temporal logic specifications},
  author={Vasile, Cristian-Ioan and Raman, Vasumathi and Karaman, Sertac},
  booktitle={2017 IEEE/RSJ International Conference on Intelligent Robots and Systems (IROS)},
  pages={3840--3847},
  year={2017},
  organization={IEEE}
}

@inproceedings{sadraddini2015robust,
  title={Robust temporal logic model predictive control},
  author={Sadraddini, Sadra and Belta, Calin},
  booktitle={2015 53rd Annual Allerton Conference on Communication, Control, and Computing (Allerton)},
  pages={772--779},
  year={2015},
  organization={IEEE}
}

@inproceedings{dawson2022robust,
  title={Robust counterexample-guided optimization for planning from differentiable temporal logic},
  author={Dawson, Charles and Fan, Chuchu},
  booktitle={2022 IEEE/RSJ International Conference on Intelligent Robots and Systems (IROS)},
  pages={7205--7212},
  year={2022},
  organization={IEEE}
}

@article{meng2023signal,
  title={Signal temporal logic neural predictive control},
  author={Meng, Yue and Fan, Chuchu},
  journal={IEEE Robotics and Automation Letters},
  year={2023},
  publisher={IEEE}
}

@inproceedings{janner2022planning,
  title={Planning with Diffusion for Flexible Behavior Synthesis},
  author={Janner, Michael and Du, Yilun and Tenenbaum, Joshua and Levine, Sergey},
  booktitle={International Conference on Machine Learning},
  pages={9902--9915},
  year={2022},
  organization={PMLR}
}

@inproceedings{zhong2023guided,
  title={Guided conditional diffusion for controllable traffic simulation},
  author={Zhong, Ziyuan and Rempe, Davis and Xu, Danfei and Chen, Yuxiao and Veer, Sushant and Che, Tong and Ray, Baishakhi and Pavone, Marco},
  booktitle={2023 IEEE international conference on robotics and automation (ICRA)},
  pages={3560--3566},
  year={2023},
  organization={IEEE}
}

@article{brohan2023rt,
  title={RT-1: Robotics Transformer for Real-World Control at Scale},
  author={Brohan, Anthony and Brown, Noah and Carbajal, Justice and Chebotar, Yevgen and Dabis, Joseph and Finn, Chelsea and Gopalakrishnan, Keerthana and Hausman, Karol and Herzog, Alexander and Hsu, Jasmine and others},
  journal={Robotics: Science and Systems XIX},
  year={2023},
  publisher={Robotics: Science and Systems Foundation}
}

@inproceedings{openx2024,
  title={Open x-embodiment: Robotic learning datasets and rt-x models: Open x-embodiment collaboration 0},
  author={O’Neill, Abby and Rehman, Abdul and Maddukuri, Abhiram and Gupta, Abhishek and Padalkar, Abhishek and Lee, Abraham and Pooley, Acorn and Gupta, Agrim and Mandlekar, Ajay and Jain, Ajinkya and others},
  booktitle={2024 IEEE International Conference on Robotics and Automation (ICRA)},
  pages={6892--6903},
  year={2024},
  organization={IEEE}
}

@inproceedings{li2024roboflamingo,
  title={Vision-language foundation models as effective robot imitators},
  author={Li, Xinghang and Liu, Minghuan and Zhang, Hanbo and Yu, Cunjun and Xu, Jie and Wu, Hongtao and Cheang, Chilam and Jing, Ya and Zhang, Weinan and Liu, Huaping and others},
  booktitle={International Conference on Learning Representations},
  volume={2024},
  pages={26703--26721},
  year={2024}
}

@article{ghosh2024octo,
  title={Octo: An open-source generalist robot policy},
  author={Team, Octo Model and Ghosh, Dibya and Walke, Homer and Pertsch, Karl and Black, Kevin and Mees, Oier and Dasari, Sudeep and Hejna, Joey and Kreiman, Tobias and Xu, Charles and others},
  journal={arXiv preprint arXiv:2405.12213},
  year={2024}
}

@article{bjorck2025groot,
  title={Gr00t n1: An open foundation model for generalist humanoid robots},
  author={Bjorck, Johan and Casta{\~n}eda, Fernando and Cherniadev, Nikita and Da, Xingye and Ding, Runyu and Fan, Linxi and Fang, Yu and Fox, Dieter and Hu, Fengyuan and Huang, Spencer and others},
  journal={arXiv preprint arXiv:2503.14734},
  year={2025}
}

@article{ye2026starvlaalpha,
  title={StarVLA-$\alpha$: Reducing Complexity in Vision-Language-Action Systems},
  author={Ye, Jinhui and Gao, Ning and Yang, Senqiao and Zheng, Jinliang and Wang, Zixuan and Chen, Yuxin and Chen, Pengguang and Chen, Yilun and Liu, Shu and Jia, Jiaya},
  journal={arXiv preprint arXiv:2604.11757},
  year={2026}
}

@inproceedings{pi2025pi05,
  title={{$\pi_{0.5}$}: a Vision-Language-Action Model with Open-World Generalization},
  author={Black, Kevin and Brown, Noah and Darpinian, James and Dhabalia, Karan and Driess, Danny and Esmail, Adnan and Equi, Michael Robert and Finn, Chelsea and Fusai, Niccolo and Galliker, Manuel Y and others},
  booktitle={9th Annual Conference on Robot Learning},
  year={2025}
}

@inproceedings{zhen2024threedvla,
  title={3D-VLA: A 3D Vision-Language-Action Generative World Model},
  author={Zhen, Haoyu and Qiu, Xiaowen and Chen, Peihao and Yang, Jincheng and Yan, Xin and Du, Yilun and Hong, Yining and Gan, Chuang},
  booktitle={International Conference on Machine Learning},
  pages={61229--61245},
  year={2024},
  organization={PMLR}
}

@article{qu2025spatialvla,
  title={Spatialvla: Exploring spatial representations for visual-language-action model},
  author={Qu, Delin and Song, Haoming and Chen, Qizhi and Yao, Yuanqi and Ye, Xinyi and Ding, Yan and Wang, Zhigang and Gu, JiaYuan and Zhao, Bin and Wang, Dong and others},
  journal={arXiv preprint arXiv:2501.15830},
  year={2025}
}

@inproceedings{feng2025diffusion,
  title={Diffusion meets options: Hierarchical generative skill composition for temporally-extended tasks},
  author={Feng, Zeyu and Luan, Hao and Ma, Kevin Yuchen and Soh, Harold},
  booktitle={2025 IEEE International Conference on Robotics and Automation (ICRA)},
  pages={10854--10860},
  year={2025},
  organization={IEEE}
}

@inproceedings{vaezipoor2021ltl2action,
  title={Ltl2action: Generalizing ltl instructions for multi-task rl},
  author={Vaezipoor, Pashootan and Li, Andrew C and Icarte, Rodrigo A Toro and Mcilraith, Sheila A},
  booktitle={International Conference on Machine Learning},
  pages={10497--10508},
  year={2021},
  organization={PMLR}
}

@article{lindemann2021funnel,
  title={Funnel control for fully actuated systems under a fragment of signal temporal logic specifications},
  author={Lindemann, Lars and Dimarogonas, Dimos V},
  journal={Nonlinear Analysis: Hybrid Systems},
  volume={39},
  pages={100973},
  year={2021},
  publisher={Elsevier}
}

@article{lindemann2019feedback,
  title={Feedback control strategies for multi-agent systems under a fragment of signal temporal logic tasks},
  author={Lindemann, Lars and Dimarogonas, Dimos V},
  journal={Automatica},
  volume={106},
  pages={284--293},
  year={2019},
  publisher={Elsevier}
}

@article{jiang2023vima,
  title={Vima: Robot manipulation with multimodal prompts},
  author={Jiang, Yunfan and Gupta, Agrim and Zhang, Zichen and Wang, Guanzhi and Dou, Yongqiang and Chen, Yanjun and Fei-Fei, Li and Anandkumar, Anima and Zhu, Yuke and Fan, Linxi},
  year={2023}
}

@inproceedings{liu2024grounding,
  title={Grounding dino: Marrying dino with grounded pre-training for open-set object detection},
  author={Liu, Shilong and Zeng, Zhaoyang and Ren, Tianhe and Li, Feng and Zhang, Hao and Yang, Jie and Jiang, Qing and Li, Chunyuan and Yang, Jianwei and Su, Hang and others},
  booktitle={European conference on computer vision},
  pages={38--55},
  year={2024},
  organization={Springer}
}

@inproceedings{jackermeier2025deepltl,
  title={Deepltl: Learning to efficiently satisfy complex ltl specifications for multi-task rl},
  author={Jackermeier, Mathias and Abate, Alessandro},
  booktitle={International Conference on Learning Representations},
  volume={2025},
  pages={14000--14028},
  year={2025}
}

\clearpage
\appendix

\twocolumn[
\begin{minipage}{\textwidth}
\centering
\vspace{-1mm}
\setlength{\tabcolsep}{5pt}
\begin{tabular}{@{}llrrrrrrr@{}}
\toprule
Domain & Method
& @1 & @16 & @64 & @128 & @256 & @512 & @1024 \\
\midrule
Panda
& Exact state
& 4.69 & 30.21 & 41.93 & 49.48 & 52.34 & 56.77 & 59.11 \\
& Visual fusion
& 3.12 & 21.61 & 38.28 & 45.31 & 50.00 & 56.25 & 62.76 \\
& Vision-TL-Action
& 3.39 & 22.66 & 39.06 & 46.61 & 52.86 & 59.64 & \textbf{67.45} \\
\midrule
AntMaze
& Exact state
& 61.46 & 89.06 & 93.75 & 94.79 & 96.88 & -- & -- \\
& Visual $4^2$
& 26.04 & 60.94 & 80.73 & 84.38 & 92.19 & -- & -- \\
& Vision-TL-Action
& 43.75 & 85.94 & 91.15 & 94.27 & \textbf{96.35} & -- & -- \\
\bottomrule
\end{tabular}
\captionof{table}{Complete three-seed mean Success@$K$ values (percent). The full
candidate curves reveal the distinction between single-sample concentration
and maximum-set support coverage.}
\label{tab:all_budgets}
\vspace{1mm}
\end{minipage}
]

\section{Complete Experimental Record}

\subsection{Evidence Partition}
We partition completed runs before drawing further conclusions. The main paper
uses only fixed-order held-out evaluations, full candidate prefixes, and the
three sampling seeds 1007, 2027, and 3407 for its planning comparisons.
Single-seed experiments are retained as controlled component diagnostics.
Earlier checkpoints, data-scale controls, and graph perturbations are reported
below as auxiliary evidence and are not pooled into the main effect. Loader
smoke tests, batch-size probes, failed launches, and attention-export-only runs
verify engineering behavior but are not counted as scientific comparisons.

\subsection{Complete Candidate-Set Values}
Table~\ref{tab:all_budgets} gives the formal values plotted in
Figure~\ref{fig:candidate_budgets} and retains the earlier AntMaze visual model
as a development reference. Each row is the mean over three sampling seeds
evaluated on the same task ordering. Candidate prefixes are nested within one
generation run, so Success@$K_1\leq K_2$ compares the same first $K_1$ samples
rather than an independently resampled set.

\FloatBarrier
\section{Representation-Resolution Diagnostics}
\label{sec:resolution_diagnostics}

\subsection{Why Native $16\times16$ Tokens?}
The target view exposes a concrete bottleneck in visual tokenization. Among 59
AntMaze tasks containing object pairs, $4\times4$ tokens cause at least one
object collision in 72.88\% of tasks and 19.42\% of object pairs. At
$8\times8$, these rates remain 54.24\% and 6.13\%. A $16\times16$ grid reduces
both rates to zero on the evaluation set. This diagnostic motivated the final
representation, but it is not itself part of the main method comparison.

\begin{figure*}[t]
\centering
\includegraphics[width=0.96\textwidth]{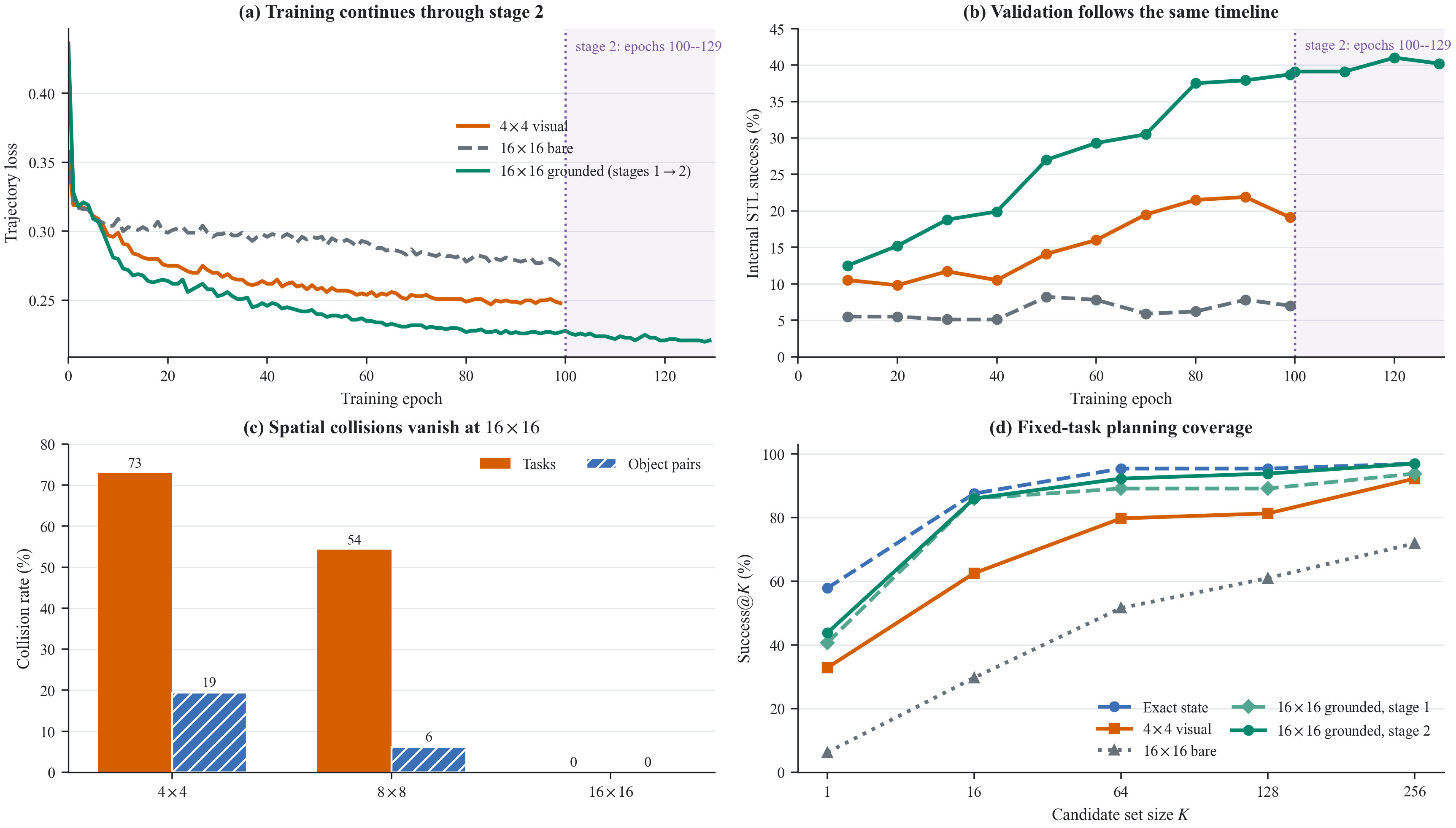}
\caption{AntMaze representation-resolution diagnostic. The first training
stage occupies epochs 0--99; the 30-epoch condition-interface adaptation is
plotted continuously at epochs 100--129 rather than restarting the horizontal
axis. Increasing spatial resolution eliminates target-token collisions, while
position/view encoding and predicate grounding convert that resolution into
planning performance.}
\label{fig:grid_resolution}
\end{figure*}

We train every $16\times16$ visual component natively rather than interpolating
or fine-tuning a $4\times4$ checkpoint. Figure~\ref{fig:grid_resolution} shows
that added resolution alone is insufficient: under the same 100-epoch schedule,
the bare model reaches only 71.88\% Success@256. Addressable position/view
tokens and predicate grounding raise coverage to 93.75\%. Thirty epochs of
limited condition-interface adaptation then reach 96.88\% on the seed-1007
gate. The final checkpoint is used; no intermediate epoch is selected post hoc.

\subsection{Resolution and Module Controls}
\begin{figure*}[t]
\centering
\includegraphics[width=0.96\textwidth]{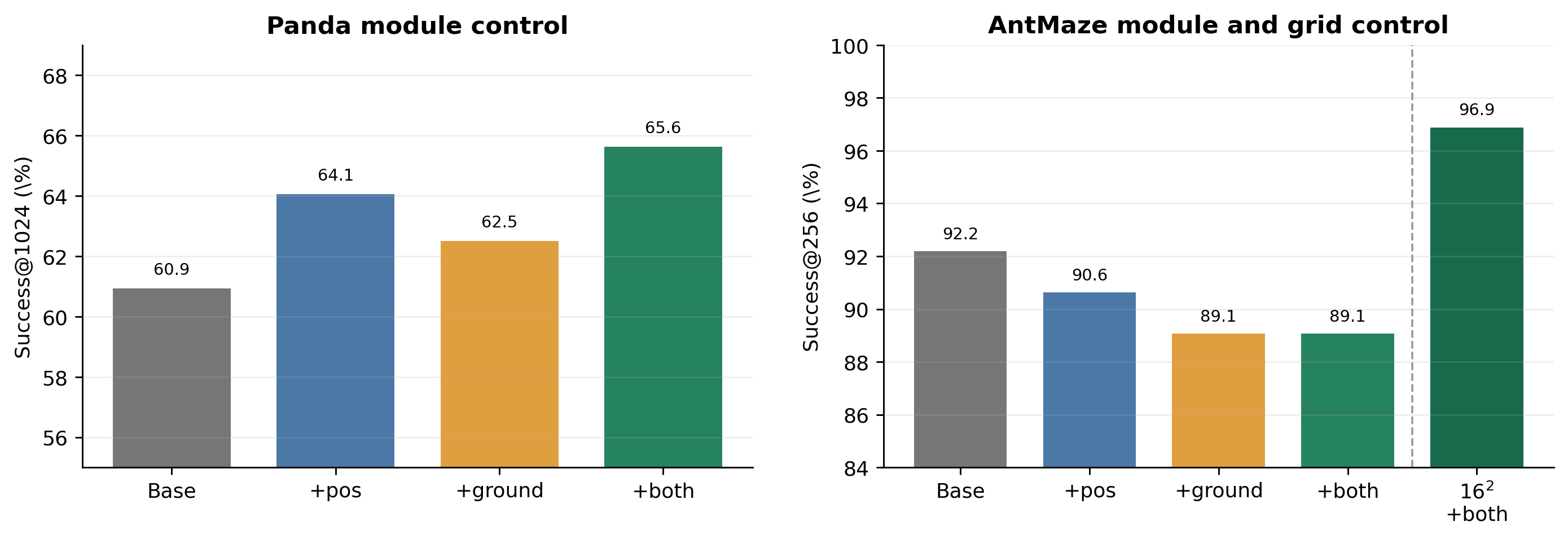}
\caption{Controlled seed-1007 module study. Position encoding and grounding
are complementary on Panda. On the collision-prone $4\times4$ AntMaze grid,
the same modules cannot recover planning performance; moving to native
$16\times16$ tokens makes their semantic supervision actionable.}
\label{fig:module_ablation}
\end{figure*}

Table~\ref{tab:ant_ablation} records the native-grid training sequence, while
Figure~\ref{fig:module_ablation} reports the complete seed-1007 $2\times2$
position and grounding gate. On Panda, position encoding raises Success@1024
from $60.94\%$ to $64.06\%$, grounding alone reaches $62.50\%$, and their
combination reaches $65.62\%$. On the collision-prone $4\times4$ AntMaze grid,
none of the three additions improves over the $92.19\%$ visual-fusion control.
The native $16\times16$ model with both additions and interface adaptation
reaches $96.88\%$. These controls show an interaction between semantic
supervision and token addressability; they are single-seed design diagnostics,
not the three-seed main-effect estimate.
\section{Task-Level and Robustness Diagnostics}
\label{sec:task_robustness}

\subsection{Performance by TL Formula Family}
Figure~\ref{fig:task_robustness}a--b decomposes the formal maximum-set results
into reach/stay, conjunction/disjunction, sequential, and until tasks. The Panda
holdout contains 11, 38, 26, and 53 tasks in these groups; AntMaze contains 5,
19, 11, and 29. On Panda, Vision-TL-Action improves most on AND/OR
($47.37\%$ versus $21.05\%$) and until ($70.44\%$ versus $61.01\%$), while
exact state remains stronger on sequential tasks ($93.59\%$ versus $76.92\%$).
On AntMaze, the visual model improves AND/OR ($96.49\%$ versus $91.23\%$),
matches reach/stay and until, and trails on sequential tasks ($87.88\%$ versus
$100\%$). The shared sequential weakness is localized rather than uniform
across formula families.

\subsection{Training-Data and Syntax-Graph Controls}
The full AntMaze split contains 39,805 trajectories; a mini control uses 503
trajectories ($1.26\%$) while preserving the 64-task evaluation. The exact-state
checkpoints reach $96.88\%$ and $39.06\%$ Success@256, a paired difference of
$+57.81$ pp with a task-bootstrap 95\% interval of $[46.35,68.75]$ pp
(Figure~\ref{fig:task_robustness}c). We therefore use the full data split in all
visual comparisons.

\begin{table}[t]
\centering
\small
\begin{tabular}{lcccc}
\toprule
AntMaze Variant & Grid & Pos. & Ground. & @256 \\
\midrule
Visual fusion & $4^2$ & No & No & 92.19 \\
Native bare & $16^2$ & No & No & 71.88 \\
Native grounded & $16^2$ & Yes & Yes & 93.75 \\
Native + interface adapt. & $16^2$ & Yes & Yes & \textbf{96.88} \\
\bottomrule
\end{tabular}
\caption{Seed-1007 AntMaze representation and interface ablation. The final
row adds 30 epochs of limited condition/time projection adaptation.}
\label{tab:ant_ablation}
\end{table}

\clearpage
\twocolumn[
\begin{minipage}{\textwidth}
\centering
\resizebox{0.94\textwidth}{!}{\includegraphics{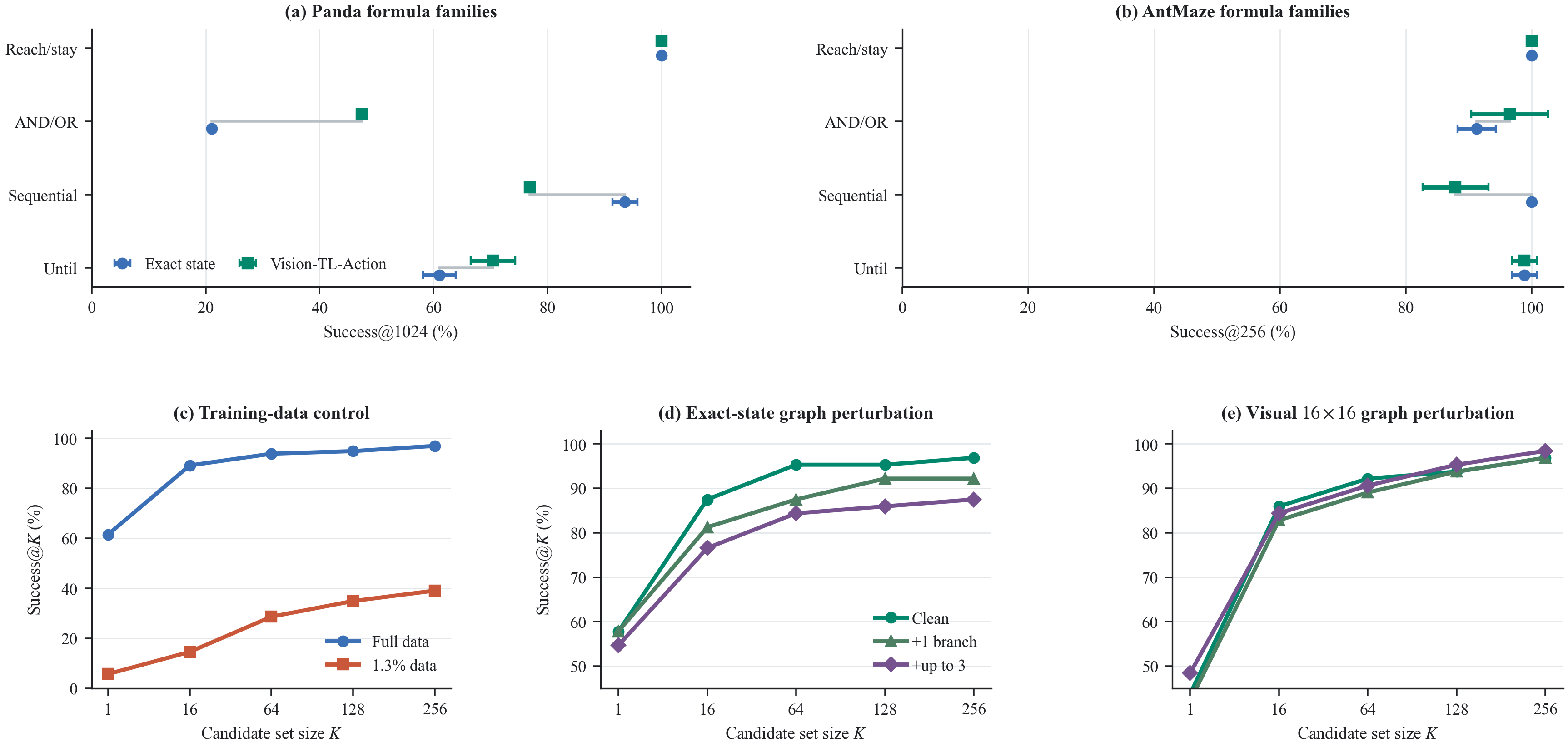}}
\captionof{figure}{Compact task and robustness diagnostics. Top: three-seed
performance by TL formula family. Bottom: training-data sufficiency and the
same redundant syntax-branch intervention applied to exact-state and final
visual checkpoints. All curves use nested candidate prefixes.}
\label{fig:task_robustness}
\vspace{1mm}
\end{minipage}
]

We also insert semantically redundant syntax branches while preserving images
and task order. At seed 1007, exact-state Success@256 decreases from $96.88\%$
on clean graphs to $92.19\%$ with one branch and $87.50\%$ with up to three.
The final $16\times16$ visual model instead obtains $96.88\%$, $96.88\%$, and
$98.44\%$ (Figure~\ref{fig:task_robustness}d--e). The clean-minus-perturbed
effects are $0.00$ pp ($[-4.69,4.69]$) and $-1.56$ pp ($[-4.69,0.00]$); the
last numerical increase is sampling variation, not evidence that redundant
syntax improves planning.

\subsection{Earlier-Checkpoint Visual-Input Audit}
Earlier checkpoints retain the qualitative visual dependence of the final
models. Panda V4 reaches $47.66\%$, $32.03\%$, $31.25\%$, and $10.16\%$ at
$K=128$ for dual, custom, bird's-eye, and zero input. The AntMaze $4\times4$
checkpoint reaches $92.19\%$, $29.69\%$, $90.62\%$, and $28.12\%$ at $K=256$
for dual, map, targets, and zero input. These values are not pooled with the
final-checkpoint effects because their models and candidate set sizes differ.

\FloatBarrier
\section{Fixed Identity-Color Grounding Audit}
\label{sec:idcolor_audit}

\subsection{Controlled Rendering and Training Protocol}
The initial visual studies left an identity ambiguity. AntMaze used six colors
modulo object ID, so O0/O6, O1/O7, and later pairs shared pixels; Panda used one
color for every target and another for every obstacle, which exposed class but
not instance identity. We therefore perform a controlled fixed-ID-color audit.
The \emph{unique} renderer assigns O0--O10 a shared, fixed 11-color palette in
both domains; O0--O5 preserve their legacy AntMaze RGB values exactly. The
\emph{legacy} renderer remains unchanged.

No expert trajectory, object placement, STL formula, action, split, or dynamics
is resampled. Panda reuses all 59,200 records and 71,325 filtered trajectories;
the rerendered index contains 112,930 record--solution keys and 225,860 image
paths, with complete training-key coverage. Legacy and unique models use the
same architecture, 32-dimensional ID embedding, task order, paired
initialization, and training/evaluation seeds (1007, 2027, and 3407). Panda is
trained for 100 epochs in both conditions; AntMaze uses the final native
$16\times16$ schedule. Success@$K$ remains oracle candidate coverage, not a
deployable top-1 selection metric.

\begin{table*}[t]
\centering
\small
\begin{tabular}{llcccc}
\toprule
Domain & Set Size & Legacy Color & Unique ID Color & Oracle State & Paired $\Delta$ [95\% CI] \\
\midrule
Panda & @1024 & $50.95\pm3.05$ & $\mathbf{63.63\pm2.86}$ & $59.11\pm1.19$ & $+12.67$ $[7.64,17.88]$ \\
AntMaze & @256 & $95.49\pm2.67$ & $\mathbf{98.09\pm1.31}$ & $96.88\pm1.56$ & $+2.60$ $[0.69,5.03]$ \\
\bottomrule
\end{tabular}
\caption{Fixed-ID-color audit at the maximum candidate set size. Values are
three-training-seed means and sample standard deviations; paired intervals
bootstrap tasks after averaging training and generation seeds.}
\label{tab:idcolor_main}
\end{table*}

\subsection{Coverage and Training-Length Effects}
\begin{figure*}[t]
\centering
\includegraphics[width=0.86\textwidth]{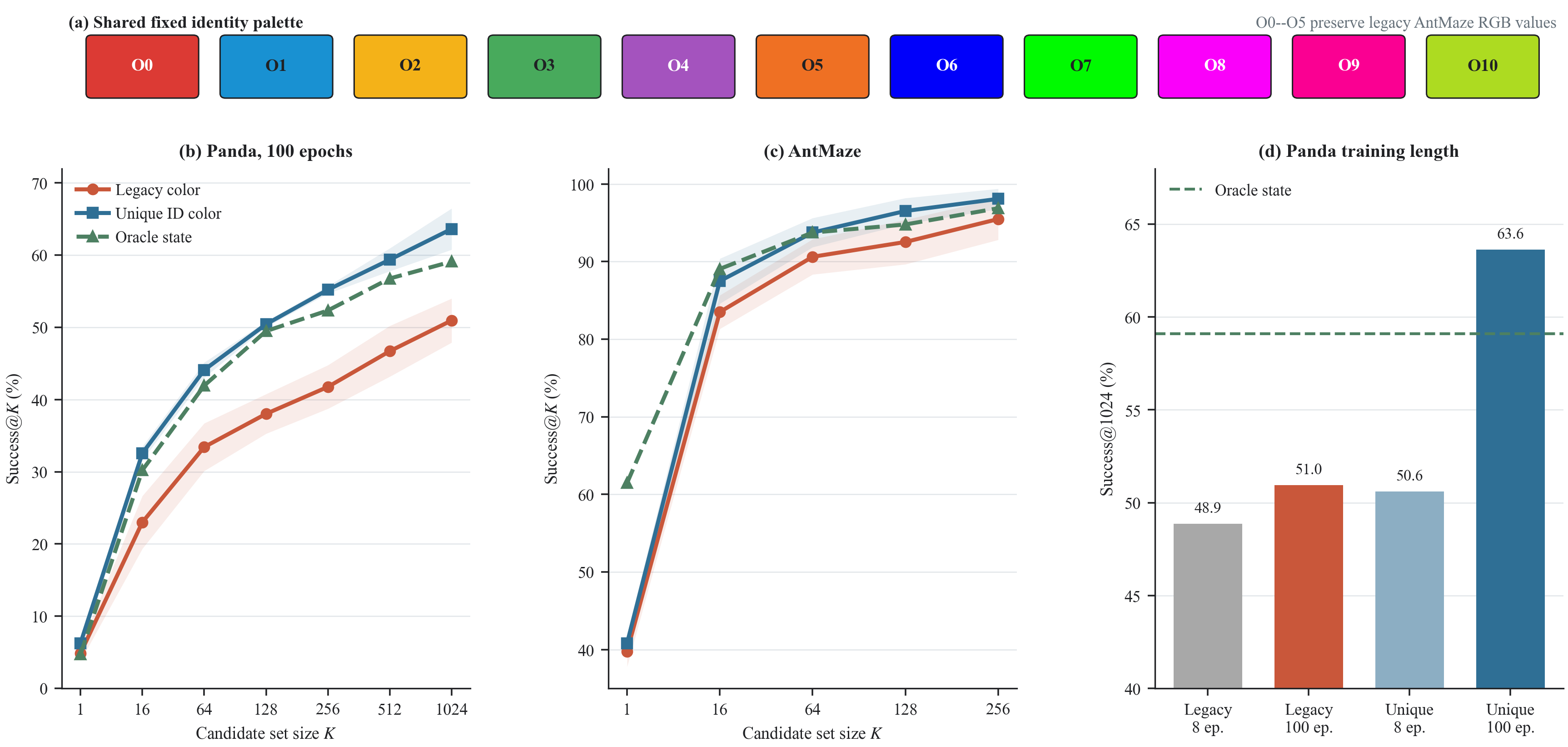}
\caption{Fixed identity-color experiment. The shared O0--O10 palette is used in
both domains. Candidate coverage compares legacy, unique, and oracle-state
models; the Panda training-length panel applies 8 and 100 epochs symmetrically.}
\label{fig:idcolor_main}
\end{figure*}

Table~\ref{tab:idcolor_main} and Figure~\ref{fig:idcolor_main} show paired gains
from changing only the color--identity relation. On Panda, the unique palette
improves Success@1024 by $12.67$ pp; on AntMaze the improvement is $2.60$ pp.
For AntMaze collision tasks, the Success@1 gain is $+6.06$ pp
($[0.51,11.62]$), versus $-1.59$ pp ($[-5.03,1.85]$) on noncollision tasks,
giving a $+7.65$ pp difference-in-differences. This same-task contrast avoids
confounding collision status with task complexity.

The Panda training-length control is equally important. Extending both methods
from 8 to 100 epochs changes Legacy Success@1024 from $48.87\%$ to $50.95\%$
($+2.08$ pp, $[-1.82,6.08]$), but changes Unique from $50.61\%$ to $63.63\%$
($+13.02$ pp, $[7.90,18.49]$). The principal identity-color effect is therefore
visible only under the longer, fair training schedule rather than by extending
one condition alone.

\subsection{Causal Interventions, Visual Controls, and Q/K Grounding}
\begin{figure*}[t]
\centering
\includegraphics[width=0.86\textwidth]{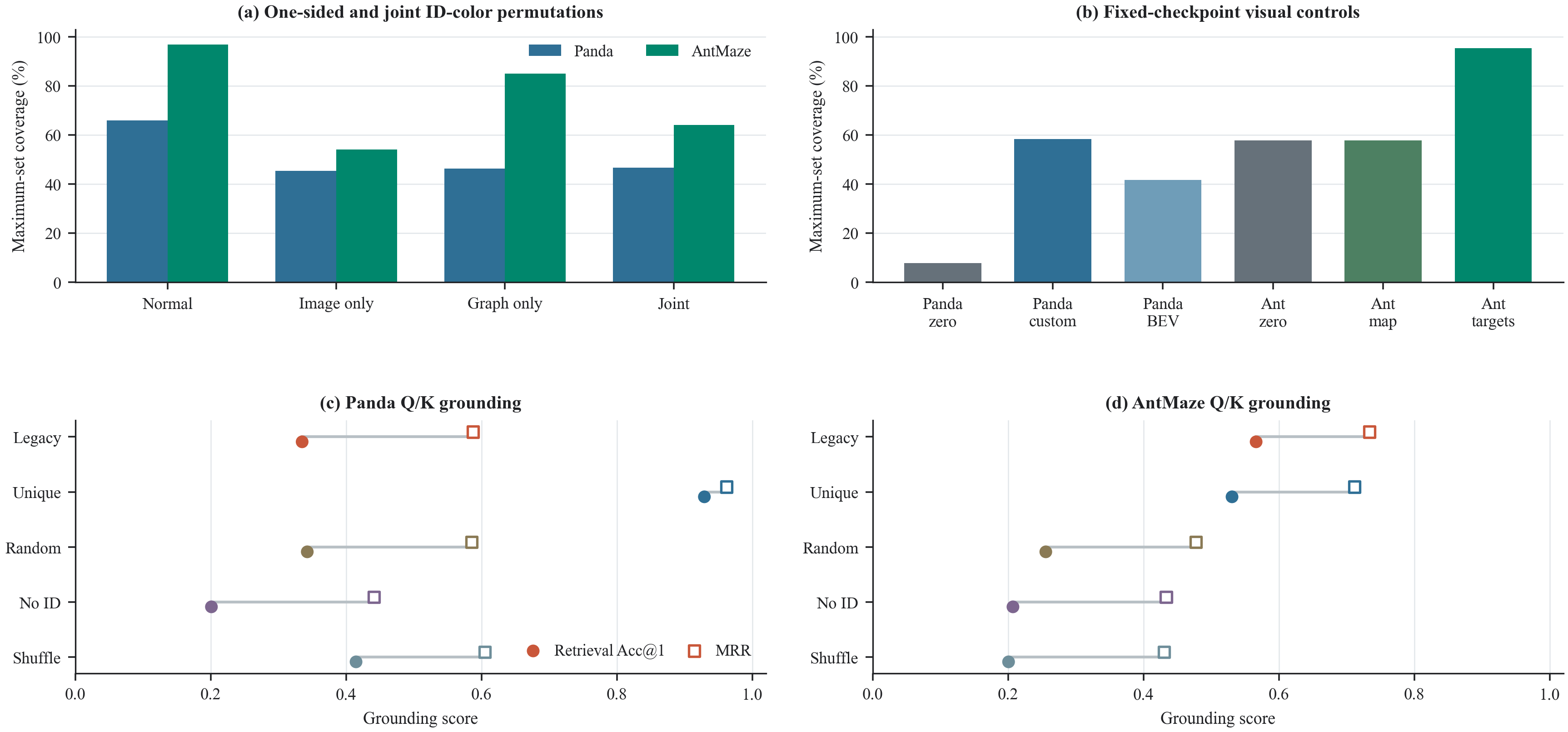}
\caption{Identity-color controls and projected-space grounding. Top:
fixed-checkpoint color/ID permutations and visual ablations. Bottom: final-layer
query-to-object-key retrieval for Legacy, Unique, RandomScene, NoID, and
ShuffleID models. Filled circles show Acc@1; open squares show MRR.}
\label{fig:idcolor_causal_qk}
\end{figure*}

We permute image colors only, graph IDs only, or both consistently while keeping
the task fixed. Maximum-set coverage falls from $65.89\%$ to $45.31\%$,
$46.35\%$, and $46.61\%$ on Panda, and from $96.88\%$ to $54.17\%$,
$84.90\%$, and $64.06\%$ on AntMaze. The one-sided drops establish sensitivity
to both modalities. The joint intervention does \emph{not} restore normal
performance, so the evidence does not establish complete permutation
equivariance. Zero-image and single-view controls in
Figure~\ref{fig:idcolor_causal_qk}b further rule out a purely nonvisual shortcut.

We compare graph queries and object-region visual keys only after their learned
cross-attention projections. In Panda, Unique raises final-layer retrieval
Acc@1 from 0.335 to 0.929 and MRR from 0.588 to 0.962, while reducing mean
localization error from 34.78 to 1.14 pixels. Random-scene, No-ID, and Shuffle-ID
controls are all substantially lower. In AntMaze, Unique improves localization
error from 4.57 to 3.66 pixels but does not improve retrieval Acc@1
(0.531 versus 0.566) or MRR (0.712 versus 0.734). Thus stable identity colors
produce strong representation alignment in Panda and a smaller, local effect in
AntMaze; they improve behavior in both domains but do not by themselves prove a
fully causal binding mechanism.

\section{Development and Training Diagnostics}

\subsection{Panda Condition Adaptation}
The internal Panda development trace starts from node-level visual fusion,
then adapts the final visual block, and finally adapts the existing
condition/time projections without changing architecture. At seed 1007, the
three stages reach $54.69\%$, $59.38\%$, and $60.94\%$ Success@1024. Because
these stages informed design selection, Figure~\ref{fig:training_diagnostics}
is a
post-hoc development record rather than independent evidence. The formal result
uses a disjoint 128-task holdout and three sampling seeds.

\subsection{Grounding Optimization Dynamics}
\begin{figure}[!ht]
\centering
\includegraphics[width=0.78\columnwidth]{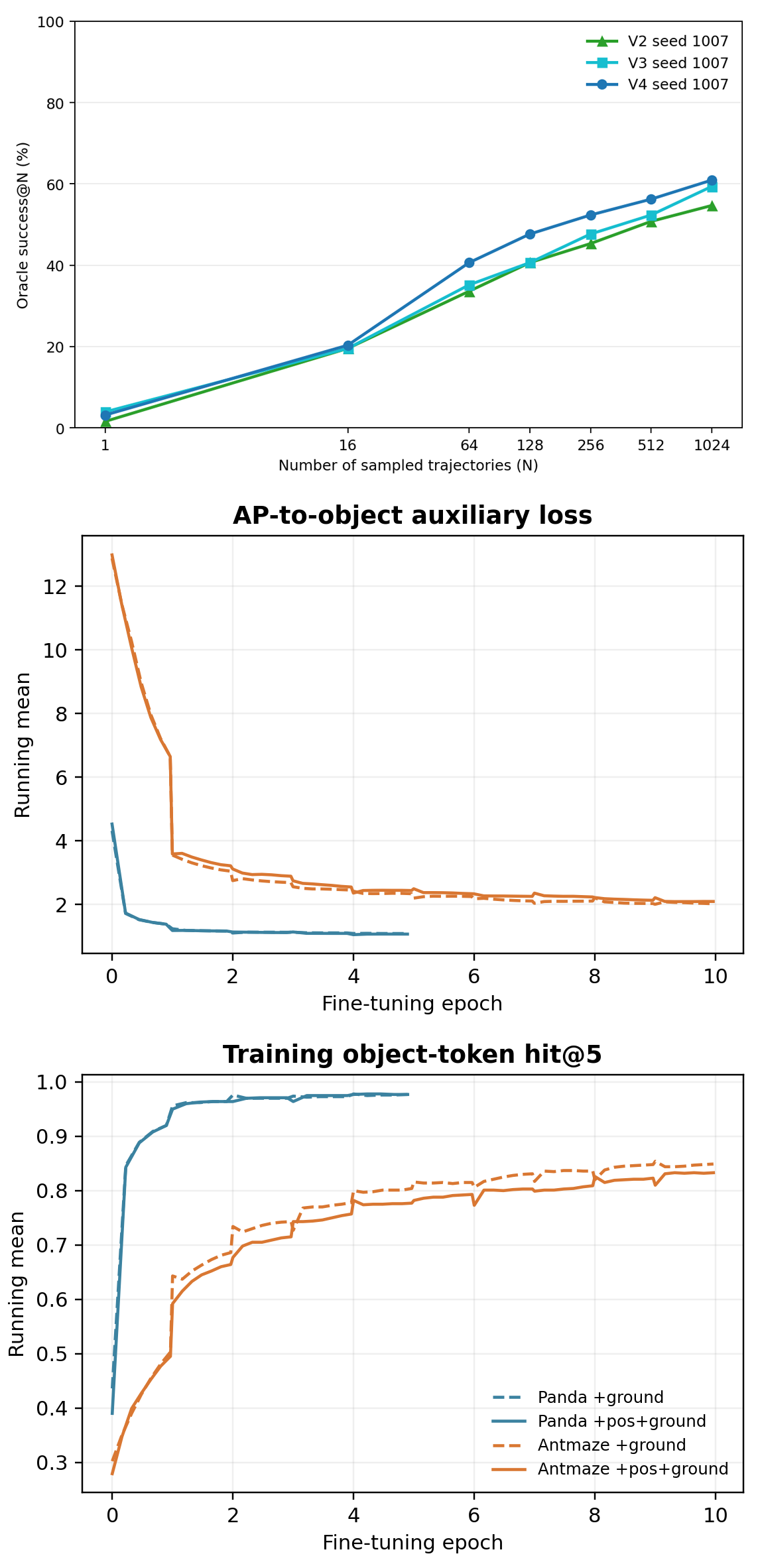}
\caption{Development and grounding diagnostics. Top: Panda internal
checkpoint progression at seed 1007; V2, V3, and V4 denote node fusion,
final-block adaptation, and condition-interface adaptation. Bottom:
grounding-stage optimization, where solid lines include spatial position
encoding and dashed lines omit it. The AntMaze curves use the earlier
$4\times4$ gate and motivate the native $16\times16$ experiment.}
\label{fig:training_diagnostics}
\end{figure}

Figure~\ref{fig:training_diagnostics} also records the auxiliary loss and
object-region hit@5 during the grounding stage. Panda hit@5 approaches $0.98$
within the
five-epoch stage. The coarse-grid AntMaze gate improves more slowly and
saturates near $0.83$--$0.85$, consistent with unresolved object-token
collisions. The loss is therefore optimizable, but a lower auxiliary loss alone
does not guarantee better planning; representation resolution and held-out
Success@$K$ remain necessary selection criteria.

\subsection{Generation Cost}
Runtime is measured per task on the same RTX 3090 machine and includes candidate
generation and robustness evaluation. At $K=1024$, Panda exact state, visual
fusion, and Vision-TL-Action take 3.261, 3.275, and 3.264 seconds per task. At
$K=256$, AntMaze exact state and Vision-TL-Action take 8.141 and 8.145 seconds.
The native AntMaze $4\times4$ and $16\times16$ training runs peak at 3.21 and
4.40--4.41 GB GPU memory, respectively. Thus the denser token grid increases
training memory, while the measured end-to-end evaluation time remains nearly
unchanged relative to exact state.

\section{Extended Grounding Evidence}
\label{sec:extended_grounding}

\subsection{Full Positive Intervention Examples}
\begin{figure*}[t]
\centering
\includegraphics[width=0.96\textwidth]{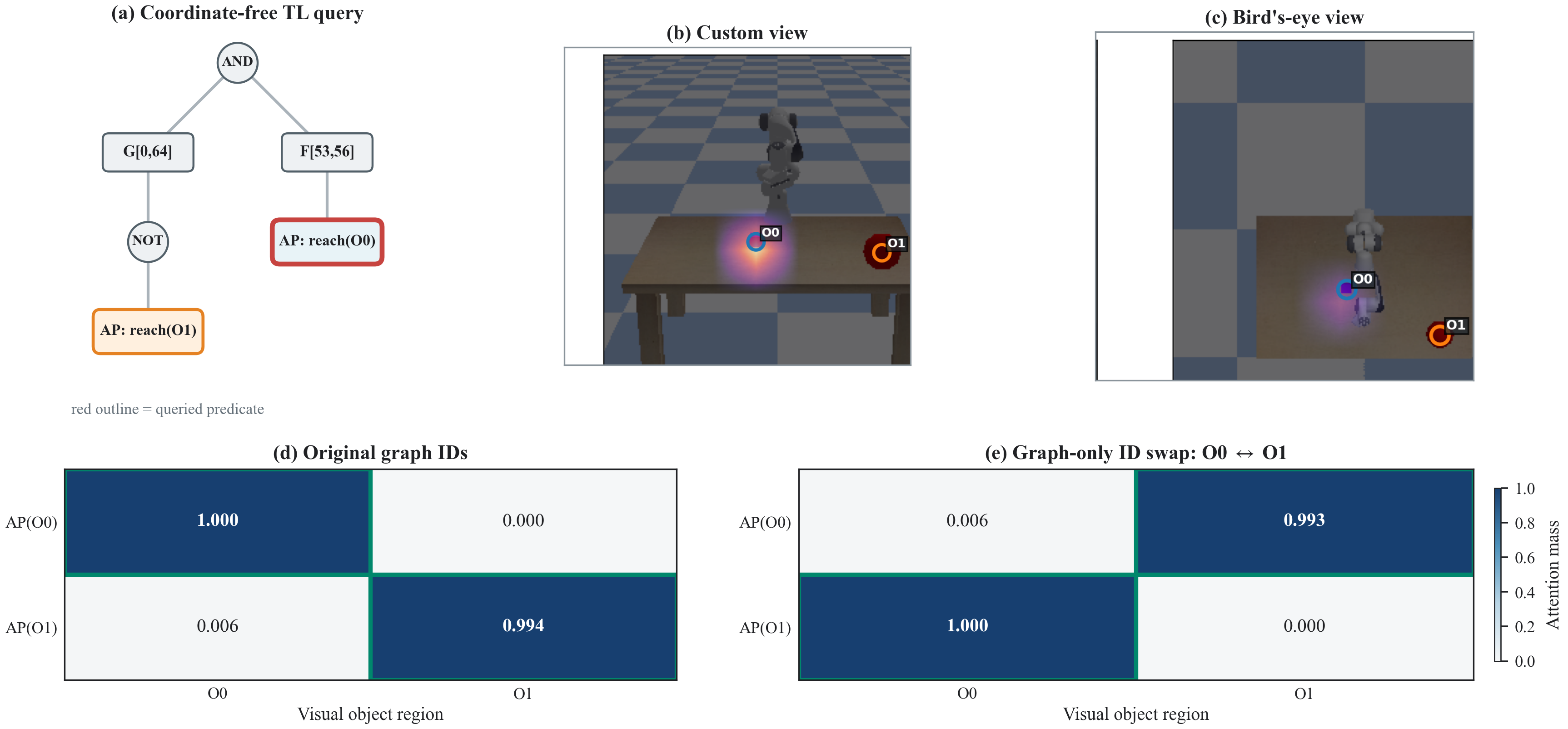}
\caption{Complete Panda positive example. The queried AP attends to its
referenced object in both camera views. Swapping only graph identities swaps the
dominant AP-to-object correspondence; outlined cells mark the expected binding.}
\label{fig:panda_ground_success_appendix}
\end{figure*}

\begin{figure*}[t]
\centering
\includegraphics[width=0.96\textwidth]{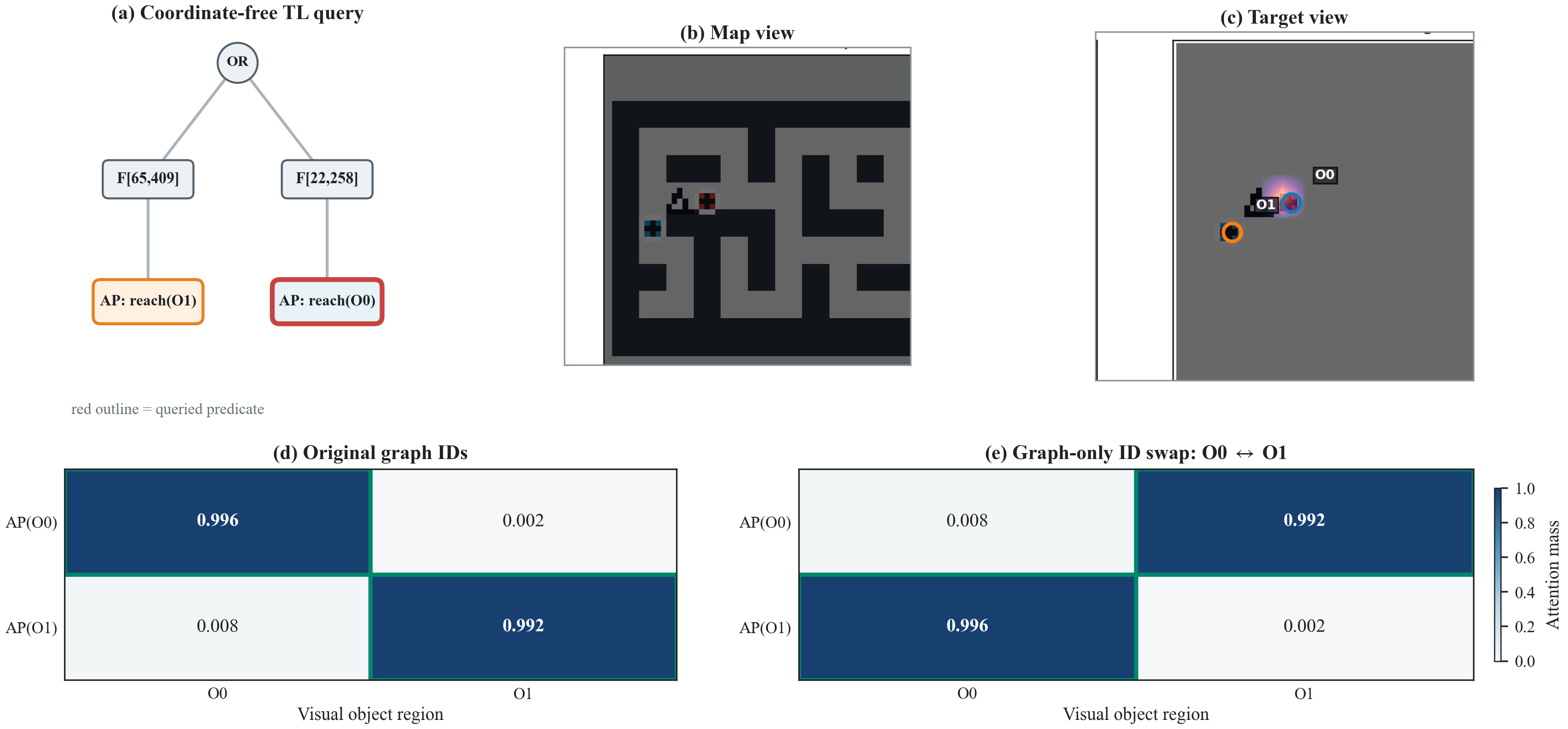}
\caption{Complete AntMaze positive example under the same graph-only identity
intervention. Both image rasters remain fixed while the high-attention entries
follow the exchanged graph identities.}
\label{fig:ant_ground_success_appendix}
\end{figure*}

Figures~\ref{fig:panda_ground_success_appendix} and
\ref{fig:ant_ground_success_appendix} expand the compact main-paper view. They
retain both visual streams, the full coordinate-free TL query, and the original
and graph-swapped AP-to-object matrices. Operator nodes carry their logical or
temporal operator and interval; AP nodes carry predicate type and object
identity, but no object coordinates. The intervention changes only these object
identities, leaving the rendered scene fixed.

\subsection{Paired Full-Task Effects}
\begin{figure}[!ht]
\centering
\includegraphics[width=0.86\columnwidth]{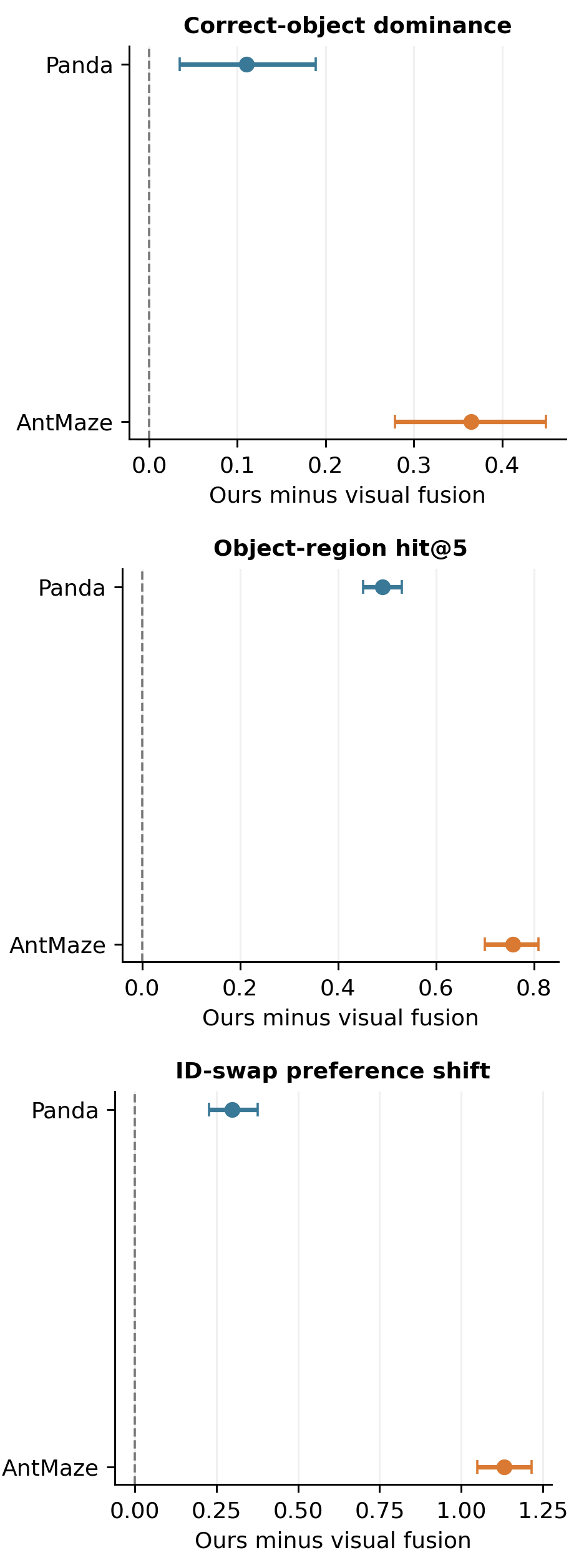}
\caption{Paired grounding improvements over visual fusion. Points are mean
task-level differences and bars are 95\% bootstrap intervals.}
\label{fig:attention_effects}
\end{figure}

Figure~\ref{fig:attention_effects} reports task-paired changes relative to each
domain's visual fusion model. On Panda, dominance, hit@5, and ID-swap shift
increase by $0.110$ ($[0.034,0.189]$), $0.491$ ($[0.451,0.531]$), and $0.298$
($[0.227,0.376]$). On AntMaze, the corresponding native-$16\times16$ effects
are $0.365$ ($[0.279,0.450]$), $0.756$ ($[0.700,0.810]$), and $1.131$
($[1.049,1.215]$). All intervals are task-bootstrap 95\% intervals. This
full-set analysis is the quantitative counterpart to the qualitative examples.
The AntMaze comparator is the earlier $4\times4$ visual-fusion checkpoint, so
its effect combines representation and grounding changes and is retained as a
diagnostic rather than a controlled main-paper ablation.

\subsection{Failure Cases Are Retained}
\begin{figure*}[t]
\centering
\includegraphics[width=0.72\textwidth]{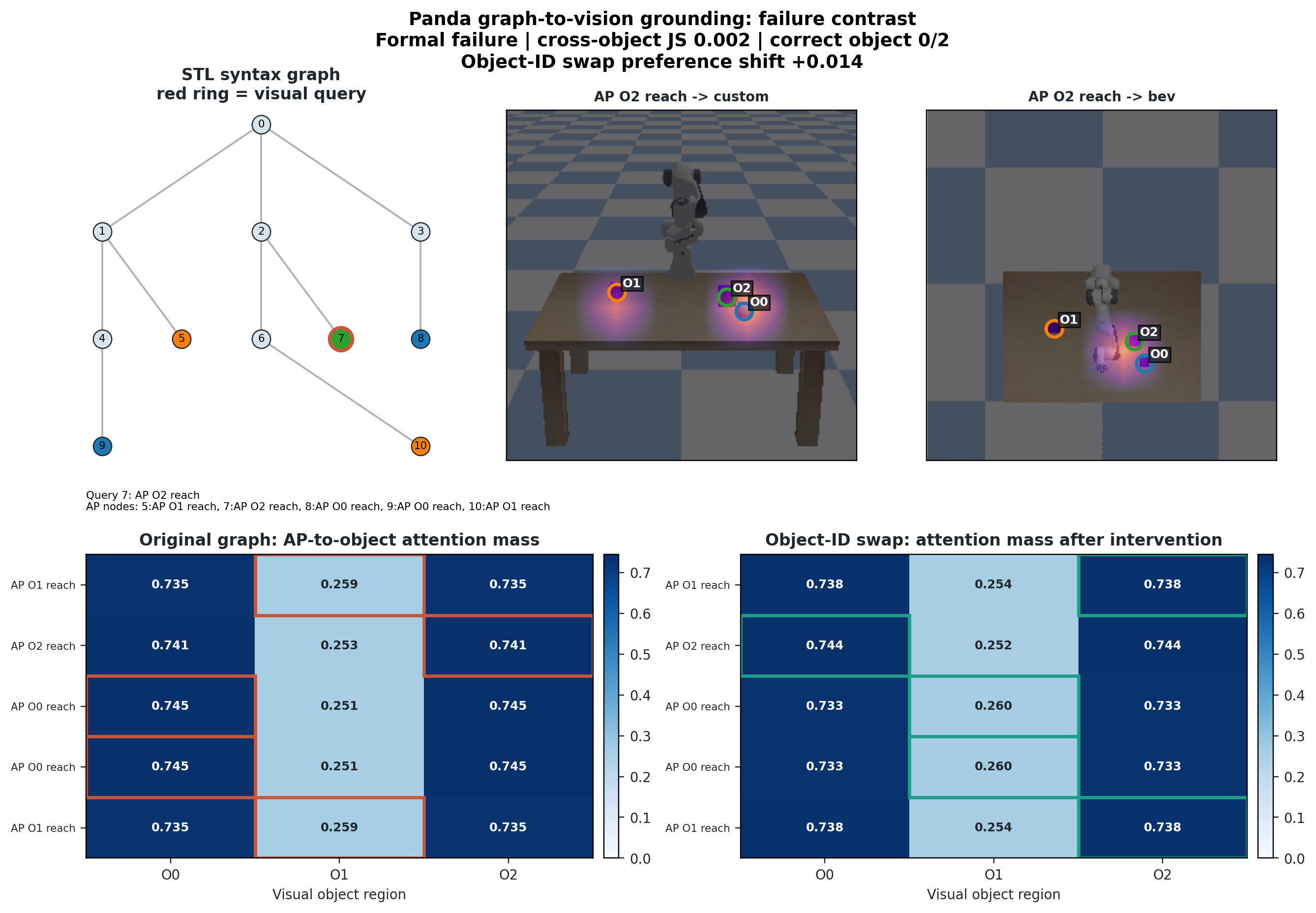}

\vspace{-1mm}
\includegraphics[width=\textwidth]{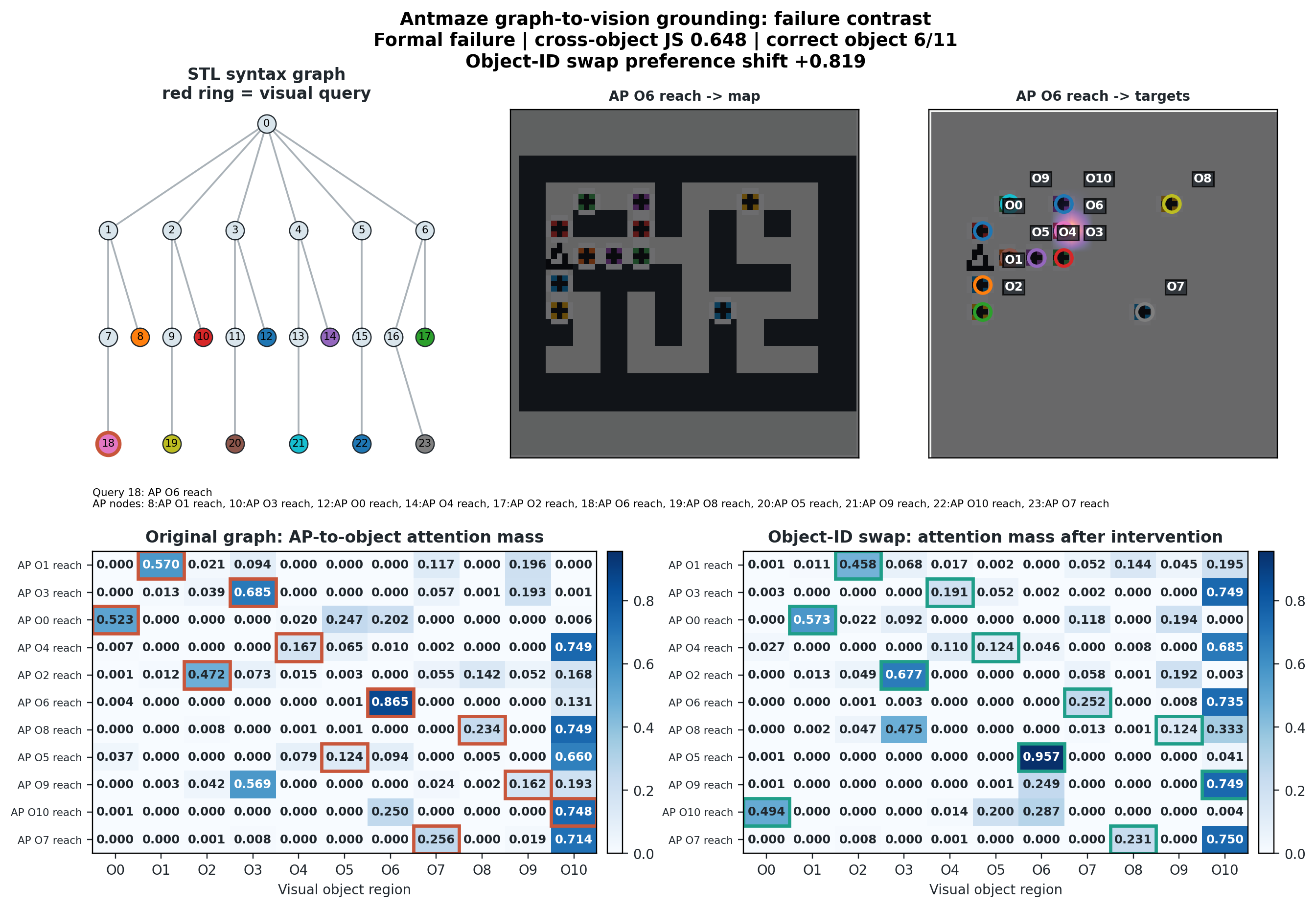}
\caption{Retained failure contrasts. Top: Panda exhibits diffuse, repeated
AP-to-object rows and a weak ID-intervention response, exposing a grounding
failure. Bottom: AntMaze binds much of the graph correctly, but its sampled
trajectory set still violates the TL task, separating grounding quality from
downstream generation quality.}
\label{fig:panda_ground_failure}
\label{fig:ant_ground_failure}
\end{figure*}

We do not select only clean attention maps. In the Panda failure of
Figure~\ref{fig:panda_ground_failure}, several AP rows attend similarly across
objects and ID swapping barely changes the preference (shift $0.014$). In the
AntMaze failure of Figure~\ref{fig:ant_ground_failure}, 6 of 11 APs dominate
their correct object regions and the swap shift remains positive ($0.819$), yet
no satisfying candidate is generated. Grounding is therefore neither perfect
nor sufficient: trajectory-generation errors remain after a predicate has
localized its object.

\section{Reproducibility Details}

\subsection{Checkpoints and Result Ownership}
We assign stable aliases Panda-PG, Panda-VF, and Panda-State to the formal
grounded, visual-fusion, and exact-state checkpoints. Ant-PG16, Ant-VF4, and
Ant-State denote the corresponding native-$16\times16$, $4\times4$, and
exact-state checkpoints. The machine-readable experiment manifest maps these
aliases to full checkpoint directories, result files, driver scripts, and
analysis entry points. Each
reported task stores its
fixed evaluation position, record ID, TL ID, formula family, candidate-prefix
successes, runtime, and sampled trajectory.

\subsection{Audit Rules}
The Panda tune and holdout partitions each contain 128 unique TL tasks and have
zero TL-ID overlap. AntMaze uses 64 unique tasks in a fixed order. Compared
methods share task keys and candidate set sizes; paired analyses abort on any key
mismatch. Main planning effects first average the three sampling seeds within a
task and then bootstrap tasks 20,000 times. Attention effects use paired task
bootstraps and 20,000 two-sided sign flips. No intermediate AntMaze epoch is
selected after inspecting the formal task set.

\newpage

\end{document}